\documentclass[journal,10pt,twoside,web]{IEEEtran}

\usepackage{cite}
\usepackage{amsmath,amssymb,amsfonts}
\usepackage{graphicx}
\usepackage{textcomp}
\usepackage{styles/style}
\usepackage[colorlinks=true, allcolors=blue]{hyperref} 
\usepackage{indentfirst}
\usepackage{rotating}
\usepackage{algorithm,algorithmic}
\usepackage{tikz}
\usepackage{tikz-3dplot}
\usetikzlibrary{arrows,shapes,automata,decorations.text,decorations.pathmorphing,backgrounds,positioning,fit}
\usepackage{enumitem}
\usepackage{color}
\usepackage{tabularx} 
\usepackage{pgfplots}
\usetikzlibrary{positioning}
\usepackage{rotating}
\usepackage{hyperref}
\usepackage{dblfloatfix}
\usepackage{url,amsmath,setspace,amssymb}
\usepackage{tikz}
\usepackage{graphicx}
\usepackage{caption}
\usepackage{subcaption}
\usepackage{float}
\usepackage{soul}
\usepackage{epsfig}
\usetikzlibrary{external}
\definecolor{royalblue}{RGB}{65, 105, 225}
\definecolor{seagreen}{RGB}{46, 139, 87}
\definecolor{firebrick}{RGB}{178,34,34}
\definecolor{darkviolet}{RGB}{138, 43, 226}
\definecolor{carrotorange}{RGB}{237, 145, 33}
\pgfplotsset{every tick label/.append style={font=\small}}
\newcommand{\mytitle}{Communication-efficient Distributed Cooperative Learning with Compressed Beliefs}

\def\BibTeX{{\rm B\kern-.05em{\sc i\kern-.025em b}\kern-.08em
    T\kern-.1667em\lower.7ex\hbox{E}\kern-.125emX}}
\title{\mytitle}

\author{
\IEEEauthorblockN{Mohammad Taha Toghani and C\'esar A. Uribe}
\\\IEEEauthorblockA{Department of Electrical and Computer Engineering, Rice University, Houston, TX, USA
\\\{\href{mailto:mttoghani@rice.edu}{mttoghani}, \href{mailto:cauribe@rice.edu}{cauribe}\}@rice.edu}
}

\begin{document}
\maketitle

\begin{abstract}
We study the problem of distributed cooperative learning, where a group of agents seeks to agree on a set of hypotheses that best describes a sequence of private observations. In the scenario where the set of hypotheses is large, we propose a belief update rule where agents share compressed (either sparse or quantized) beliefs with an arbitrary positive compression rate. Our algorithm leverages a unified communication rule that enables agents to access wide-ranging compression operators as black-box modules. We prove the almost sure asymptotic exponential convergence of beliefs around the set of optimal hypotheses. Additionally, we show a non-asymptotic, explicit, and linear concentration rate in probability of the beliefs on the optimal hypothesis set. We provide numerical experiments to illustrate the communication benefits of our method. The simulation results show that the number of transmitted bits can be reduced to $5-10\%$ of the non-compressed method in the studied scenarios.
\end{abstract}


\section{Introduction}
\label{sec:introduction}
During the past decade, the analysis of distributed systems has seen a dramatic rise in interest. Fundamental limitations and structural properties of distributed systems such as limited memory, communication bandwidth, and lack of a central coordinator require coordination by distributed information sharing. Analysis of social networks as well as sensor networks~\cite{jadbabaie2012non,olfati2006belief,Halme2019BayesianMH,Nedic2017DistributedLF}, distributed inference~\cite{Chen2019ResilientDE,AlSayed2017RobustDE,Nassif2019DistributedIO}, and multi-agent control~\cite{Nedic2009DistributedSM,Olshevsky2017LinearTA} are applications of distributed learning.

We consider the problem of decision-making in a network, where agents observe a stream of private signals and exchange their beliefs to agree on hypothesis set that best describes their observations. Fully Bayesian solutions require agents to have a complete knowledge of the whole distributed system, such as each other's likelihood functions~\cite{gale2003bayesian,acemoglu2011bayesian}. On the other hand, locally Bayesian or non-Bayesian methods~\cite{lalitha2018social,Molavi2018ATO,hare2020non} alternatively suggest agents (i) update their beliefs internally using the Bayes rule and (ii) combine their beliefs locally among neighbors using a fusion rule.

A broad line of research has been developed to address different aspects of social/cooperative learning~\cite{lalitha2017learning,uribe2019non,ntemos2021social}, among which efficient communication is one of the most fundamental challenges~\cite{matta2019exponential,salhab2020social,sundaram2020distributed}. In such decentralized problems, agents are restricted only to access the information from their local neighbors. Several works investigated non-Bayesian learning from various perspectives for fixed undirected networks. Likewise, other communication setups such as directed, non-connected, and time-varying networks~\cite{nedic2015nonasymptotic,salami2017social,matta2020interplay}, as well as adaptive update rules\cite{ying2016information,bordignon2021adaptation} have been explored thoroughly.

Conventional non-Bayesian algorithms require agents to share their beliefs on all hypotheses with their neighbors~\cite{Nedic2017DistributedLF,Molavi2018ATO,nedic2017fast}. This, however, imposes large communication loads should the set of hypotheses be large. Similarly, the idea of quantized communication has been extensively studied previously~\cite{seide20141,karimireddy2019error,lin2020achieving,stich2018sparsified}. Several methods have been recently proposed to address the communication bottlenecks over the networks~\cite{doan2020fast,kovalev2020linearly,koloskova2019decentralized,koloskova2019decentralized2,song2021compressed}. In social learning moreover, it may not be crucial that agents exchange all their beliefs with each other; instead, a set of compressed messages could be shared through the network.
Works in~\cite{Sundaram2020DistributedHT} and~\cite{mitra2020distributed} propose algorithms with a compressed message sharing with the assumption of a unique common parameter locally optimal for all agents. Furthermore, these algorithms consider unweighted mixing matrices for communication besides specific sparsification and quantization methods. More importantly, no non-asymptotic analysis is available even under those stronger assumptions. In this paper, we work with milder assumptions such as weighted networks and conflicting hypotheses, i.e., the set of parameters that best describes all agents' observations (on average) may not be locally optimal for all agents. We furthermore seek to provide the first non-asymptotic analysis for non-Bayesian learning with compressed communication. Besides, our algorithm provides a unified framework that accommodates a wide range of compression operators (Section~\ref{sec:setup}). In~\cite{Bordignon2020SocialLW}, authors study the possibility of answering binary questions about a particular hypothesis by sending a subset of beliefs. They propose an algorithm with partial information sharing to reduce communication. In contrast, we will present a more general approach that contains various quantization and sparsification operators.

This paper proposes a distributed non-Bayesian algorithm for social learning where agents exchange their compressed beliefs. The core of our algorithm is inspired by CHOCO-GOSSIP~\cite{koloskova2019decentralized}, but we develop a modified version of their results to show convergence in our algorithm. Our proposed algorithm inherits CHOCO's benefits like arbitrary compression rate and mild assumptions on the quantizer.

In summary, the contributions of this paper are threefold:
\begin{itemize}
    \item We propose a novel algorithm for non-Bayesian distributed learning with (possibly) arbitrary compressed communication per round. Our algorithm follows a unified consensus mechanism covering various compression operators, including sparsification and quantization functions. Thus, it provides a simple and general communication rule for agents to leverage a proper compression operator as a black-box module. We also show a memory-efficient version of the algorithm.
    \item We provide a non-asymptotic, explicit, and linear convergence rate of beliefs for our algorithm in probability. We work under the conflicting hypotheses setup, where optimal hypotheses of each agent locally need not be the optimal hypotheses of the network. We also prove exponential asymptotic convergence of the beliefs around the set of optimal hypotheses almost surely.
    \item Finally, we show the communication advantages of our algorithm through numerical experiments on various compression operators and multiple network topologies.
\end{itemize}

The remainder of this paper is organized as follows. In Section~\ref{sec:setup}, we describe the problem setup and state our main algorithm and results. In Section~\ref{sec:asymptotic-analysis}, we prove the almost sure asymptotic exponential convergence rate for the proposed algorithm. Likewise, we provide the non-asymptotic convergence rate in probability for our algorithm in Section~\ref{sec:non-asymptotic-analysis}.
In Section~\ref{sec:experiments}, we illustrate the proposed algorithm via numerical experiments. Finally, we end with concluding remarks and discussing future works in Section~\ref{sec:conclusion}.

\vspace{1em}

\textbf{Notation:} We write $[n]$ to denote the set \mbox{$\{1,\dots,n\}$}. We use the notation of bolding for vectors and matrices. For a matrix \mbox{$\mA\in\bbR^{n\times n}$}, we write \mbox{$[\mA]_{ij}$} or \mbox{$\mA_{ij}$} to denote the entry in the $i$-th row and $j$-th column. We use \mbox{$\mI_n$} for the identity matrix of size \mbox{$n\times n$} as well as \mbox{$\vect{1}_n$} for the vector of all one with size $n$ where we drop the subscript for brevity. We refer to agents and time by subscripts and superscripts, respectively. We write \mbox{$\lambda_i(\mA)$} to denote the $i$-th eigenvalue of matrix $\mA$ in terms of magnitude where \mbox{$|\lambda_1(\mA)|\geq|\lambda_2(\mA)|\geq\dots\geq|\lambda_n(\mA)|$}. For arbitrary vectors \mbox{$\vx,\vy\in\bbR^n$}, \mbox{$\log\vx$}, \mbox{${\vx}.{\vy}$}, and \mbox{${\vx}/{\vy}$} denote element-wise $\log$, product, and division, respectively. \mbox{$\lVert\vx\rVert$} denotes \mbox{$2$-norm} of vector $\vx$. 

\section{Problem Setup And Main Results}
\label{sec:setup}

We begin by describing the communication network structure, observation model, and distributed non-Bayesian method for the social learning problem. We state the corresponding optimization formulation and discuss its properties. We then explain the properties of our desired compression operators and introduce some examples that satisfy those properties. We further present our algorithm and provide a detailed explanation for it. Finally, we present our two main theorems (asymptotic and non-asymptotic convergence) along with the assumption that guarantees the underlying results.

\vspace{1em}

\textbf{Communication Network Structure:}
We consider a group of $n$ agents interacting over a fixed, undirected, and connected communication network \mbox{$\mcG = \{[n],\mcE\}$} where \mbox{$\mcE\subset [n]\times [n]$} is the set of edges. If there exists an edge between agent $i$ and agent $j$, they may communicate to each other, i.e., send and receive information through the corresponding link. We also consider an arbitrary doubly stochastic and symmetric mixing matrix $\mA$ compliant with network $\mcG$ such that \mbox{$\mA_{ij}=0$} for \mbox{$(i,j)\notin\mcE$} with positive diagonal entries, thus \mbox{$1=\lambda_1(\mA)>|\lambda_2(\mA)|$}. Let \mbox{$\delta \triangleq 1-|\lambda_2(\mA)|$} denote the spectral gap of \mbox{$\mA$} which is in interval \mbox{$(0,1]$}. Given the description of the network, we next formally describe the observation model along with the social learning problem and discuss its formal objective function.

\vspace{1em}

\textbf{Social Learning Problem:}
Let each agent \mbox{$i\in[n]$} observe a sequence of i.i.d. random variables \mbox{$S^1_i,S^2_i,\dots$} with realizations \mbox{$s^1_i,s^2_i,\dots$} over the course of time. The underlying random variables \mbox{$S^t_i$}, for all \mbox{$t\geq 1$}, take values in some measurable space \mbox{$(\mcS_i,\mcA_i)$}, where \mbox{$\mcS_i$} is the realization space and \mbox{$\mcA_i$} is the corresponding \mbox{$\sigma$-algebra}. We assume that random variables \mbox{$\{S_i^t\}$} are independent and identically distributed according to a common unknown distribution (measure) $f_i$ on \mbox{$(\mcS_i,\mcA_i)$}, i.e., $S_i^t \sim f_i$, for all $t$. Let \mbox{$\Theta=\{\theta_1,\theta_2,\dots,\theta_m\}$} denote a set of $m$ parameters (hypotheses) which in social learning is referred to as the possible \textit{states of the world}. We also assume that each agent $i$ knows a likelihood statistical model \mbox{$\{\ell_i\left(.|\theta_k\right)\}_{k=1}^{m}$}, conditional on the set of parameters $\Theta$, where \mbox{$\ell_i\left(.|\theta_k\right)$} is a probability distribution on \mbox{$\mcS_i$}, for all $k\in[m]$. We additionally assume that the probability distributions are independent across the agents. In this setup, we do not require that the existence of a single (unique) \mbox{$\theta\in\Theta$} such that \mbox{$\ell_i\left(.|\theta\right)=f_i(.)$} almost everywhere, for all $i\in[n]$. For an agent $i\in[n]$ and a hypothesis \mbox{$\theta\in\Theta$}, we denote \mbox{$\bbP_i^\theta = \bigotimes_{t=1}^{\infty} \ell_i\left(.|\theta\right)$} as the probability measure of the sequence of observed signals \mbox{$\left(s^1_i,s^2_i,\dots\right)$}. We therefore use \mbox{$\bbE_\theta[.]$} to denote the joint expectation operator associated with measures \mbox{$\bbP_1^\theta,\bbP_2^\theta,\dots,\bbP_n^\theta$}. We now describe the objective function and then introduce the distributed non-Bayesian method.

Given the described setup above, we now formally state the set optimal hypotheses and the corresponding objective function for this problem. The set of agents collaboratively try to solve the following optimization problem:
\begin{align}\label{eq:non-bay-optim}
\Theta^\star = \argmin_{\theta\in\Theta} F(\theta) \triangleq   \frac{1}{n}\sum_{i=1}^{n} D_\mathrm{KL} \left(f_i\|\ell_i\left(.|\theta\right)\right),
\end{align}
where \mbox{$D_\mathrm{KL} \left(f_i\|\ell_i\left(.|\theta\right)\right)$} is Kullback-Leibler (KL) divergence between the unknown distribution $f_i$ and the conditional distribution \mbox{$\ell_i\left(.|\theta\right)$}, $\Theta^\star$ is the set of optimal parameters.
We denote \mbox{$F^\star$} to be the optimal objective value, i.e., \mbox{$F^\star = F(\theta_w)$} for an arbitrary \mbox{$\theta_w\in\Theta^\star$}. The objective in Eq.~\eqref{eq:non-bay-optim} incorporates the scenario with conflicting hypotheses and does not impose the existence of any $\theta$ such that \mbox{$\ell_i\left(.|\theta\right)=f_i$}, for all $i\in[n]$. In other words, the above optimization problem returns a subset of \mbox{$\Theta$} that best describes the unknown distributions $f_i$, on average. This optimization problem does not require any optimal parameter \mbox{$\theta_w\in\Theta^\star$} to be also locally optimal for all agents \mbox{$i\in[n]$}, thus some level of resiliency to noisy scenarios (see~\cite[Fig.~7]{nedic2017fast}). We refer the reader to~\cite{jadbabaie2012non,nedic2017fast} for the comparison between the distributed and centralized frameworks.

We now provide an explanation of the distributed non-Bayesian learning approach using the concept of beliefs.

\vspace{1em}

\textbf{Distributed Non-Bayesian Learning:}
Each agent \mbox{$i\in[n]$} starts with a prior probability distribution, namely the set of prior \textit{beliefs} \mbox{$\tilde\vmu_i^0 = [\tilde\mu_i^0(\theta_1), \dots, \tilde\mu_i^0(\theta_m)]^\top$}, on the finite measurable space \mbox{$(\Theta,2^\Theta)$}. At round $(t+1)$, for \mbox{$t\geq 0$}, each agent $i$ updates its probability measure \mbox{$\tilde\vmu_i^{t+1}$}, based on (i) its recently observed signal \mbox{$s_i^{t+1}$}, and (ii) (some information of) the probability distributions \mbox{$\tilde\vmu_j^t$}, for all $j\in[n]$ such that \mbox{$(i,j)\in\mcE$}, using a Bayesian-type update rule. We refer to such methods as non-Bayesian or locally Bayesian methods. See~\cite{jadbabaie2012non,Sundaram2020DistributedHT,nedic2017fast} for more detailed descriptions. 

We hitherto introduced the objective function and gave an outline of distributed non-Bayesian learning. Before presenting a detailed explanation of our algorithm, we first describe our interesting class of compression operators.

\vspace{1em}

\textbf{Compression Operator:}
In classical non-Bayesian frameworks, agents usually exchange their beliefs of dimension $m$ at each round. Thus, given $b$ bits baseline for a floating-point scalar, agent $i$ has to send $mb$ bits to each of its neighbors for a single message. Our algorithm proposes compressing messages with a proper compression operator under a feedback-error pipeline before communication. Here, we first introduce our desired class of compression operators, and later we will discuss our algorithm using it.

Inspired by works in~\cite{alistarh2017qsgd,stich2018sparsified,koloskova2019decentralized}, we consider a class of randomized compression operators (potentially biased) \mbox{$Q:\bbR^m \times \mcZ \times (0,1] \rightarrow \bbR^m$}, that satisfy
\begin{align}\label{eq:q-comp}
    \bbE_{\vzeta}\left[\left\lVert Q(\vx,\vzeta,\omega) - \vx \right\rVert^2\right] \leq (1-\omega) \left\lVert \vx \right\rVert^2, \quad\,\, \forall \vx\in \bbR^m,
\end{align}
where \mbox{$\omega \in (0,1]$} is the desired \textit{compression ratio}, \mbox{$\vzeta$} is a random variable with output space $\mcZ$, and \mbox{$\bbE_{\vzeta}[.]$} indicates the expectation over the internal randomness of the operator. To be more specific, assume that $(\mcZ,\mcF)$ is some measurable space from which $Q(.)$ takes samples according to a uniform measure on the realizations. Also, note that realization $\vzeta$ called by $Q(.)$ are independent across the time and agents. In other words, assuming a sequence \mbox{$\{\vx_i^t\}$} for each agent \mbox{$i\in[n]$}, and at any time \mbox{$t\geq 0$}, the operator calls \mbox{$Q(\vx_i^t,{\vzeta}_i^t,\omega)$}, where \mbox{$\vzeta_i^t$} are all independent from each other. From here onwards, we drop \mbox{$\vzeta,\omega$}, and write \mbox{$Q(\vx)$} for brevity. With an abuse of the notation, for a matrix \mbox{$\mX=[\vx_1,\vx_2,\dots,\vx_n]^\top$} of size $n\times m$, we denote \mbox{$Q(\mX)=[Q(\vx_1),Q(\vx_2),\dots,Q(\vx_n)]^\top$} as the row-wise compressed function with independent \mbox{$\vzeta_1,\vzeta_2,\dots,\vzeta_n$} and a common corresponding \mbox{$\omega$}. A wide range of compression operators, containing both sparsification and quantization functions, satisfy Eq.~\eqref{eq:q-comp}, including:

\begin{itemize}
\item \mbox{$\mathrm{rand}_{k}$ or $\mathrm{rand}_{100\omega\%}$}: Randomly selecting $k$ out of $m$ coordinates and setting the rest to zero.
\item \mbox{$\mathrm{top}_{k}$ or $\mathrm{top}_{100\omega\%}$}: Selecting $k$ out of $m$ coordinates with highest magnitude and setting the rest to zero.
\item \mbox{$\mathrm{qsgd}_{k\,\mathrm{bits}}$}: Rounding each coordinate of \mbox{${\lvert\vx\rvert}/{\lVert\vx\rVert}$} to one of the \mbox{$u=2^{k-1}-1$} quantization levels or zero (\mbox{$k-1$} bits), and one bit for the sign of the coordinate. The quantization operator is defined as follows:
\begingroup
\allowdisplaybreaks
\begin{align}
    Q\left(\vx\right) &= \mathrm{qsgd}_{k\,\mathrm{bits}}\left(\vx\right) = \frac{\omega\sign(\vx).\lVert\vx\rVert}{u}\left\lfloor u\frac{\lvert\vx\rvert}{\lVert\vx\rVert}+\vzeta\right\rfloor,\nonumber\\
    \text{with}\quad\omega&=\left(1+\min\left\{{m}/{u^2},{\sqrt{m}}/{u}\right\}\right)^{-1},\label{eq:qsgd}
\end{align}
\endgroup
where \mbox{$\vzeta{\sim}[0,1]^m$} uniformly at random, and all operators ($\sign(.)$, \mbox{$\lfloor.\rfloor$}, and products) are element-wise.
\end{itemize}
Table~\ref{tab:compression} summarizes the compression ratio $\omega$ and the number of bits required for encoding an $m$-dimensional message with each of the above three operators. For a more comprehensive list of operators (biased or unbiased), see~\cite{beznosikov2020biased,kovalev2020linearly}.

\begin{remark}[Deterministic Compression]\label{rem:q-comp-det}
Compression operators such as \mbox{$\mathrm{top}_k$} and deterministic \mbox{$\mathrm{qsgd}_{k\,\mathrm{bits}}$}\footnote{Deterministic quantization of each entry to its closest quantization level.}, provide a unique mapping given fixed $\vx, \omega$. Thus, the following inequality holds for the class of deterministic operators (cf. Eq.~\eqref{eq:q-comp}):
\begin{align}\label{eq:q-comp-det}
    \left\lVert Q(\vx) - \vx \right\rVert^2 \leq (1-\omega) \left\lVert \vx \right\rVert^2, \qquad \forall \vx\in \bbR^m.
\end{align}
\end{remark}

We are now ready to explain our algorithm, \textit{distributed non-Bayesian learning with compressed communication}.

\vspace{1em}

\begin{table}[!t]
\caption{Some example compression operators satisfying Eq.~\eqref{eq:q-comp} and their encoding bits, given
$b$ bits baseline for floating-point scalars, and $m$-dimensional vectors. For \mbox{$\mathrm{rand}_k$} and \mbox{$\mathrm{top}_k$}, \mbox{$k\in[m]$} indicates the number of coordinates (out of $m$) being transferred. For \mbox{$\mathrm{qsgd}_{k\,\mathrm{bits}}$}, \mbox{$k\in[b]$} indicates the number of bits by which we quantize each coordinate.}
\centering
\begin{tabular}{l|lll}
$Q$ operator & type& $\omega$   & encoding bits\\
\hline\hline
$\mathrm{rand}_{k}$ & sparsification & ${k}/{m}$ & $k\left(b+\log m\right)$\\
$\mathrm{top}_{k}$& sparsification & ${k}/{m}$ & $k\left(b+\log m\right)$ \\
$\mathrm{qsgd}_{k\,\mathrm{bits}}$  & quantization& Eq.~\eqref{eq:qsgd}  & $mk+b$ \\
\hline
$\mathrm{full}$& no compression& 1 & mb
\end{tabular}
\label{tab:compression}
\end{table}

\textbf{Main Algorithm:}
Agent $i$ begins with some prior beliefs \mbox{$\tilde\vmu_i^0$}, then at each round \mbox{$t+1$} for all \mbox{$t\geq 0$}, observing a new signal and exchanging (some information of) its beliefs with the neighbors, seeks to update its beliefs using a non-Bayesian rule. We also denoted \mbox{$\tilde\vmu_i^t$} as agent $i$'s beliefs (probability measure) at time $t$. We furthermore denote \mbox{$\vmu_i^t$} and \mbox{$\hat\vmu_i^t$}, respectively as beliefs without normalization and their approximation, where \mbox{$\vmu_i^0=\tilde\vmu_i^0$}, and  \mbox{$\hat\vmu_i^0=\vect{1_m}$}. Note that unlike $\tilde\vmu_i^t$ which is a probability measure, the elements of \mbox{$\vmu_i^t$} or \mbox{$\hat\vmu_i^t$} do not add up to one. Classical methods for distributed non-Bayesian learning require the agents to communicate all $m$ beliefs with their neighbors~\cite{jadbabaie2012non,nedic2017fast}. We however propose an error-feedback scheme inspired by~\cite{koloskova2019decentralized,stich2018sparsified,koloskova2019decentralized2}, where in each agent \mbox{$i\in[n]$} transmits some compressed version of the ratio between \mbox{$\vmu_i^t$} and \mbox{$\hat\vmu_i^t$} to its neighbors, at round $(t+1)$ (for \mbox{$t\geq 0$}). This way, agent $i$'s neighbors can internally recover \mbox{$\hat{\vmu}_i^{t+1}$}, i.e., the approximation of \mbox{$\vmu_i^{t}$}, and use it to update their \mbox{$\vmu_j^{t+1}$} by a Bayesian rule. Our method is thoroughly described in Algorithm~\ref{alg:compressed-social-learning} wherein Steps~\ref{ln:q-comp},~\ref{ln:mu-hat}, and~\ref{ln:mu} respectively specify the compression procedure, approximation update, and Bayesian update. Through Steps~\mbox{\ref{ln:q-comp}-\ref{ln:end-for}}, agent $i$ and its neighbors modify their estimation of agent $i$'s beliefs. One can simply see that as of Step~\ref{ln:q-comp}, the communication can be reduced per round. Note that in our proposed algorithm, the agents interact with each other synchronously (i.e., in parallel). We emphasize this fact, in the corresponding step of the pseudo-code.

\begin{figure}[t]
\centering
\includegraphics[width=0.8\linewidth]{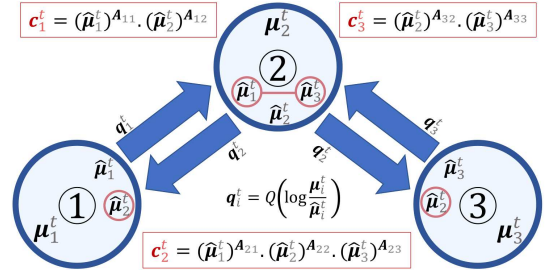}
\caption{A group of three agents communicating over a path network. The parameters required in the local memory of each node are shown around it. In accordance with step~\ref{ln:mu-hat} of Algorithm~\ref{alg:compressed-social-learning}, each node keeps an approximation of its neighbors' beliefs. A memory-efficient variation could be obtained by keeping a geometric average of neighbors' parameters.}
\label{fig:network-path-3-mu}
\end{figure}

We now summarize our algorithm with the following update rule: for all $i\in[n]$, and $\theta\in\Theta$,
\begingroup
\allowdisplaybreaks
\begin{subequations}\label{eq:update-mu-alg}
\begin{align}
    \hat{\mu}_i^{t+1}(\theta) &= \hat{\mu}_i^{t}(\theta) \exp\left\{Q^{\theta}\left(\log\left({\vmu_i^{t}}/{\hat{\vmu}_i^{t}}\right)\right)\right\},\label{eq:update-mu-1}\\
    \mu_i^{t+1}(\theta) &= \mu_i^{t}(\theta) \displaystyle\prod_{j=1}^{n}\left(\frac{\hat{\mu}_j^{t+1}(\theta)}{\hat{\mu}_i^{t+1}(\theta)}\right)^{\gamma\mA_{ij}}.\ell_{i}(s_i^{t+1}|\theta),\label{eq:update-mu-2}\\
    \tilde{\mu}_i^{t+1}(\theta) &= {\mu_i^{t+1}(\theta)} \Big/{\displaystyle\sum_{k=1}^{m}\mu_i^{t+1}(\theta_k)},\label{eq:update-mu-3}
\end{align}
\end{subequations}
\endgroup
where $\gamma\in(0,1]$ is the learning stepsize, and $Q^\theta$ indicates the entry corresponding to $\theta$ (recall that the output of $Q(.)$ is a vector). The value of $\gamma$ depends on the network topology $\delta$ and the compression ratio $\omega$. Figure~\ref{fig:network-path-3-mu} for example portrays the interaction between $3$ agents on a path network. In Section~\ref{sec:asymptotic-analysis}, we will discuss how our algorithm, i.e., Eq.~\eqref{eq:update-mu-alg}, turns into a variation of~\cite[Eq.~(2)]{nedic2017fast} without normalization, when $\omega=1$.

\begin{algorithm}[t]
    \caption{Distributed Non-Bayesian Learning with Compressed Communication}
	\label{alg:compressed-social-learning}
	\textbf{Input:} initial beliefs $\tilde\vmu_i^0\in\bbR^m$, mixing matrix $\mA$, compression ratio $\omega\in (0,1]$, and learning stepsize $\gamma\in (0,1]$ \\
	\textbf{Procedure :}
	\begin{algorithmic}[1]
	    \STATE{initialize $\hat{\vmu}_i^0 := \vect{1}_m$, and $\vmu_i^0 := \tilde\vmu_i^0, \quad \text{for all } i\in [n]$}
	    \FOR{$t$ \textbf{in} $0,\dots,T-1$, in parallel for all $i \in [n]$}
	    \STATE{$\vq_i^t:= Q(\log\vmu^{t}_i-\log\hat{\vmu}^{t}_i)$}\label{ln:q-comp}
	    \FOR{$j\in[n]$ such that $\mA_{ij}>0$ (including $j=i$)}
	    \STATE{Send $\vq_i^t$ and receive $\vq_j^t$}
	    \STATE{for all $\theta\in \Theta$: \\\quad$\hat{\mu}_j^{t+1}(\theta) = \hat{\mu}_j^{t}(\theta). \exp\left(q_j^t(\theta)\right)$}\label{ln:mu-hat}
	    \ENDFOR\label{ln:end-for}
	    \STATE{Observe $s_i^{t+1}$}
	    \STATE{for all $\theta\in \Theta$:
	    \\\quad$\mu^{t+1}_i(\theta) = \mu^t_i(\theta)\displaystyle\prod_{j=1}^{n} \left(\frac{\hat{\mu}^{t+1}_j(\theta)}{\hat{\mu}^{t+1}_i(\theta)}\right)^{\gamma\mA_{ij}}.\ell_i\left(s_i^{t+1}|\theta\right)$}\label{ln:mu}
	    \STATE{$\tilde{\vmu}^{t+1}_i = \frac{1}{\vect{1}^\top\vmu^{t+1}_i}\vmu^{t+1}_i$}
	    \ENDFOR
	\end{algorithmic}
	\textbf{Output:} final beliefs $\tilde{\vmu}_i^T\in\bbR^m, \quad \text{for all } i\in[n]$
\end{algorithm}

\textbf{Memory-Efficient Algorithm:} Algorithm~\ref{alg:compressed-social-learning} requires all agents to keep the approximation of their neighbors' beliefs locally. So, it initially seems that the memory requirements for each node are proportional to its degree. However, a simple memory-efficient implementation similar to~\cite[Appendix~E]{koloskova2019decentralized} can be employed to address this problem. Specifically, each agent $i$, can keep three vectors ($3m$ parameters) $\vmu_i^t$, $\hat{\vmu}_i^t$, and $\vc_i^t=\prod_{j:\mA_{ij}>0}\left(\hat{\vmu}_i^t\right)^{\mA_{ij}}$ where the power and product operators are entrywise. Figure~\ref{fig:network-path-3-mu} illustrates which parameters should be replaced with $\vc_i^t$, for the path example three agents. The memory-efficient pseudo-code is presented in Appendix~\ref{app:memory-efficient-alg}.

Next, we state our assumptions on the initial conditioning and conditional likelihood distributions. This will guarantee the convergence properties of our algorithm.

\begin{assumption}\label{assump:init}
For all agents $i\in[n]$ and parameters $\theta\in\Theta$, the following properties hold:
\begin{enumerate}[label=(\alph*)]
    \item $\exists\alpha_1>0$ such that $\tilde\mu_i^0(\theta)>\alpha_1$,\label{assump:init-mu}
    \item $\exists\alpha_2>0$ such that if $f_i(s_i^t)>0$ then $\ell_i\left(s_i^t|\theta\right)>\alpha_2$.\label{assump:init-ell}
\end{enumerate}
\end{assumption}

For instance, uniform prior beliefs on all hypotheses, \mbox{$\tilde\vmu_i^0=\frac{1}{m}\vect{1}_m$}, satisfy Assumption~\ref{assump:init}\ref{assump:init-mu}, and is reasonable when there is no prior information.
Furthermore, Assumption~\ref{assump:init}\ref{assump:init-ell} implies a lower bound on the likelihood measures. This assumption has been studied before and it is common in the literature~\cite{nedic2017fast,uribe2019non,boucheron2013concentration,ntemos2021social}.

We now state our first result on asymptotic convergence of the beliefs generated by Eq.~\eqref{eq:update-mu-alg}. We indeed show that the logarithm of the ratio between any non-optimal and optimal beliefs is strictly decreasing. This, in turn, implies the concentration of beliefs around the set of optimal hypotheses, almost surely.

\begin{theorem}[Asymptotic Convergence]\label{thm:asymptotic}
Let Assumption~\ref{assump:init} holds and the compression operator $Q(.)$ satisfies Eq.~\eqref{eq:q-comp}. There exists $\gamma>0$, such that $\tilde\mu_i^t(\theta)$ generated by Eq.~\eqref{eq:update-mu-alg}, has the following property: for all \mbox{$i\in[n]$}, \mbox{$\theta_v\notin\Theta^\star$}, \mbox{$\theta_w\in\Theta^\star$},
\begin{align*}
    \lim_{t\to\infty}\frac{1}{t}\bbE_{\vzeta}\left[\log\frac{\tilde\mu_i^{t}(\theta_v)}{\tilde\mu_i^{t}(\theta_w)}\right]=-C_v\quad \mathrm{a.s.},
\end{align*}
where \mbox{$C_v\triangleq F(\theta_v)-F^\star$} is a positive constant.
\end{theorem}

A detailed proof of Theorem~\ref{thm:asymptotic} is presented in Section~\ref{sec:asymptotic-analysis}. 
Note that two sources of randomness play roles in the above result, (i) the almost sure convergence refers to the randomness associated with the observation spaces $(\mcS_i,\mcA_i)$ and likelihood measures $\bbP_i^\theta$, while (ii) the expectation $\bbE_{\vzeta}[.]$ indicates the randomness originated by the compression operator. The results of Theorem~\ref{thm:asymptotic} imply that asymptotically, the randomness of the beliefs will depend only on the internal randomness of the compression operator, not on the randomness of the observations.  Next we state Corollary~\ref{cor:asymptotic-det}, where we specify the asymptotic result for a class of deterministic operators (see Remark~\ref{rem:q-comp-det}) and show that normalized beliefs $\tilde{\mu}^t_i(\theta_v)$ for any non-optimal hypotheses $\theta_v\notin\Theta^\star$ converges to zero, almost surely.

\begin{corollary}[Asymptotic Convergence: Deterministic Compression Operator]\label{cor:asymptotic-det}
Let Assumption~\ref{assump:init} holds and compression operator $Q(.)$ satisfies Eq.~\eqref{eq:q-comp-det}. If there exists only one optimal belief, i.e., $\Theta^\star=\{\theta^\star\}$, then with the same $\gamma$ as Theorem~\ref{thm:asymptotic},
\begin{align*}
    \lim_{t\to\infty}\tilde\mu_i^{t}(\theta^\star)=1\quad \mathrm{a.s.}\quad \text{for all } \mbox{$i\in[n]$}\text{, and }\mbox{$\theta^\star\in\Theta^\star$}.
\end{align*}
\end{corollary}

Theorem~\ref{thm:asymptotic} and Corollary~\ref{cor:asymptotic-det} state asymptotic convergence guarantees. We now present our theoretical result regarding the explicit non-asymptotic convergence rate.

\begin{theorem}[Non-Asymptotic Convergence Rate]\label{thm:non-asymptotic}
Let Assumption~\ref{assump:init} holds, the compression operator $Q(.)$ satisfies Eq.~\eqref{eq:q-comp}, and set \mbox{$\gamma\triangleq  {\delta^2\omega}/({32\delta+2\delta^2+8\beta^2+4\delta\beta^2-8\delta\omega})$}. For an arbitrary probability \mbox{$\rho\in(0,1)$}, \mbox{$\tilde{\mu}_i^t(\theta)$} generated by Eq.~\eqref{eq:update-mu-alg} has the following property:
there is some integer \mbox{$T(\rho)\geq 0$} such that, with probability at least \mbox{$1-\rho$},
\begin{align*}
    \bbE_{\vzeta}\left[\log{\tilde{\mu}}_i^{t}(\theta_v)\right]\leq -\frac{t}{2}C_1 + C_2 \quad\text{for all } \mbox{$i\in[n]$}\text{, and }\mbox{$\theta_v\notin\Theta^\star$},
\end{align*}
for all $t\geq T(\rho)$, where
\begingroup
\allowdisplaybreaks
\begin{align*}
    C_1 &\triangleq \min_{\substack{\theta_v\notin\Theta^\star}}\left(F(\theta_v)-F^\star\right), \, \, C_2 \triangleq \frac{162\sqrt{nm}}{\delta^2\gamma\omega}\log\frac{1}{\alpha},\\
    & \text{and} \,\,\, T(\rho) \triangleq \frac{8}{C_1^2}\left(\log\alpha\right)^2\log\frac{1}{\rho},
\end{align*}
\endgroup
with \mbox{$\alpha=\min\{\alpha_1,\alpha_2\}$}, and \mbox{$\beta = \left\lVert\mI-\mA\right\rVert_2$}.
\end{theorem}
We will present the proof for Theorem~\ref{thm:non-asymptotic} in Section~\ref{sec:non-asymptotic-analysis}. Before stating the analysis, we first discuss the non-asymptotic result for the case of deterministic compression operator in Corollary~\ref{cor:non-asymptotic-det}, and next, seek to construe the convergence rate.

\begin{corollary}[Non-Asymptotic Convergence Rate: Deterministic Compression Operator] \label{cor:non-asymptotic-det}
Under the same scenario as Theorem~\ref{thm:non-asymptotic}, with the additional assumption that the compression operator $Q(.)$ satisfies Eq.~\eqref{eq:q-comp-det}, then with probability at least \mbox{$1-\rho$}, the following property holds:
\begin{align*}
    \tilde{\mu}_i^{t}(\theta_v)\leq \exp\left(-\frac{t}{2}C_1 + C_2\right) \quad \text{for all } \mbox{$i\in[n]$}\text{, and }\mbox{$\theta_v\notin\Theta^\star$},
\end{align*}
with the same $C_1$, $C_2$, and $T(\rho)$ as in Theorem~\ref{thm:non-asymptotic}.
\end{corollary}

We now briefly discuss the convergence of Corollary~\ref{cor:non-asymptotic-det}.

\textbf{Convergence Rate Interpretation:} Corollary~\ref{cor:non-asymptotic-det} suggests an explicit linear rate in terms of $t$, providing a probabilistic upper-bound on the beliefs outside the optimal set. This implies that with a high probability, \mbox{$\tilde\mu_i^t(\theta_v)$} exponentially vanishes to zero, for all \mbox{$i\in[n]$}, and any non-optimal hypothesis $\theta_v\notin\Theta^\star$. Formally, for an arbitrary $\epsilon>0$, with probability at least $1-\rho$, we have \mbox{$\tilde\mu_i^t(\theta_v)\leq \epsilon$}, after at least
\begin{align}\label{eq:prob-convergence-rate}
    t\geq \max\left\{\frac{2}{C_1}\left(C_2+\ln\frac{1}{\epsilon}\right), T(\rho)\right\},
\end{align}
iterations (rounds). First of all, note that parameters $C_1$ and $T(\rho)$ are the same as that of~\cite{nedic2017fast}, where $C_1$ is proportional to the difference between the first and second optimal values of function $F$ in Eq.~\eqref{eq:non-bay-optim}. Second of all, the constant parameter $C_2$ in~\cite{nedic2017fast} is of $\mcO\big(\delta^{-1}\log n\big)$. 

In our algorithm, we however know that the constant parameter $C_2$ depends on the choice of learning stepsize \mbox{$\gamma$}. According to the suboptimal choice \mbox{$\gamma=\mcO(\delta^2\omega)$} in Theorem~\ref{thm:non-asymptotic}, we can infer that \mbox{$C_2=\mcO\big(\delta^{-4}\omega^{-2}n^{\frac{1}{2}}m^{\frac{1}{2}}\big)$} in our analysis. Therefore, due to the fact that $T(\rho)$ is also proportional to $C_1^{-2}$, in scenarios with small enough $C_1$, the number of rounds required for our algorithm to reach an \mbox{$\epsilon$-convergence} can be roughly the same as that of~\cite{nedic2017fast}. Thus, in setups with small enough $C_1$ (e.g., inversely proportional to constant $C_2$ in our analysis, $C_1=\mcO\big(\frac{\delta^{4}\omega^{2}}{\sqrt{nm}}\big)$), our algorithm can save a significant number of bits.

One should keep in mind that $C_1$, $T(\rho)$, and particularly $C_2$, $\gamma$, all take conservative values. The suboptimal choice of $\gamma$ in Theorem~\ref{thm:non-asymptotic}, is indeed borrowed from~\cite{koloskova2019decentralized} that guarantees the convergence of the algorithm for any $\omega$, while one can modify the stepsize given certain scenarios. We will see in Section~\ref{sec:experiments} that our algorithm converges fast in practice while reduces the communication significantly.

The following two sections are respectively dedicated to the almost sure asymptotic and probabilistic non-asymptotic proofs.

\section{Asymptotic Convergence Analysis}\label{sec:asymptotic-analysis}

In this section, we seek to prove Theorem~\ref{thm:asymptotic}. We therefore consider virtual parameters $\vnu_i^t,\hat\vnu_i^t$, with an update rule similar to $\vmu_i^t,\hat\vmu_i^t$, but without any compression, i.e., $\omega=1$. Our proof technique is to show that $\mu_i^t(\theta)$ converges to the same value as of $\nu_i^t(\theta)$. Initially, consider the following update rule
\begingroup
\allowdisplaybreaks
\begin{subequations}\label{eq:update-nu-alg}
\begin{align}
    \hat{\nu}_i^{t+1}(\theta) &= \nu_i^{t}(\theta)\label{eq:update-nu-1},\\
    \nu_i^{t+1}(\theta) &= \nu_i^{t}(\theta) \displaystyle\prod_{j=1}^{n}\left(\frac{\hat{\nu}_j^{t+1}(\theta)}{\hat{\nu}_i^{t+1}(\theta)}\right)^{\gamma\mA_{ij}}.\ell_{i}(s_i^{t+1}|\theta),\label{eq:update-nu-2}
\end{align}
\end{subequations}
\endgroup
which is an update rule without compression, i.e., \mbox{$\omega=1$}. Eq.~\eqref{eq:update-nu-1} implies that $\hat\nu_i^{t+1}$ which is supposed to be the approximation of the most recent parameter $\nu_i^{t}$, is exactly retrieved by agent $i$'s neighbors, or equivalently \mbox{$\omega=1$}. Moreover, given the fact that $\mA$ is row-stochastic with positive diagonal entries, and due to Eq.~\eqref{eq:update-nu-1}, the update rule in Eq.~\eqref{eq:update-nu-2} can be simplified as follows:
\begin{align}\label{eq:update-nu-equiv}
    \nu_i^{t+1}(\theta) &=  \displaystyle\prod_{j=1}^{n}\nu_j^{t}(\theta)^{\mB_{ij}}\ell_{i}(s_i^{t+1}|\theta),
\end{align}
where \mbox{$\mB \triangleq (1-\gamma)\mI + \gamma\mA$} is also a doubly stochastic matrix with spectral gap \mbox{$\gamma\delta$}. By construction, we initialize \mbox{$\nu_i^0(\theta)=\mu_i^0(\theta)$}, for all \mbox{$i\in[n]$} and \mbox{$\theta\in\Theta$}. Therefore, following the proof for~\cite[Theorem~1]{nedic2017fast} (without normalization), we can conclude that, for all \mbox{$i\in[n],\theta_v\notin\Theta^\star,\theta_w\in\Theta^\star$},
\begin{align}\label{eq:optimal-solution-ratio}
    \lim_{t\to\infty}\frac{1}{t}\log\frac{\nu_i^{t}(\theta_v)}{\nu_i^{t}(\theta_w)}=-C_v\quad \mathrm{a.s.},
\end{align}
with the same $C_v$ as in Theorem~\ref{thm:asymptotic}. It is worth mentioning the update rule in~\cite{nedic2017fast} also requires a normalization step, but all of their proofs is agnostic to the normalization value, as they focus on the ratio between the beliefs.
Further, let define \mbox{$\xi_i^t(\theta)=\log\left({\nu_i^{t}(\theta)}/{\nu_i^{t-1}(\theta)}\right)$} as the ratio of each belief for two subsequent steps. As we mentioned earlier $\vxi_i^t$ indicates the vector of \mbox{$[\xi_i^t(\theta_1),\dots,\xi_i^t(\theta_m)]^\top$}. The following lemma provides an upper bound on $\lVert\vxi_i^t\rVert$.

\begin{lemma}[Bounded ratio]\label{lem:bounded-xi}
Let Assumption~\ref{assump:init} holds and \mbox{$\alpha = \min\{\alpha_1,\alpha_2\}$}. The $\nu_i^t(\theta)$ generated by Eq.~\eqref{eq:update-nu-equiv} has the following property:
\begin{align*}
    \sum_{i=1}^n\lVert\vxi_i^t\rVert^2<R^2,\quad t=1,2,\dots,
\end{align*}
where \mbox{$R\triangleq{4\sqrt{nm}}\log({1}/{\alpha})/({\gamma\delta})$}.
\end{lemma}

The proof of Lemma~\ref{lem:bounded-xi}, can be found in Appendix~\ref{app:bounded-xi},
We further state an auxiliary result that will help us prove the compressed communication property of our algorithm. Consider an update rule over $\vx_i^t,\hat{\vx}_i^t\in \bbR^m$ with the presence of some exogenous noise $\vxi_i^t\in \bbR^m$
\begingroup
\allowdisplaybreaks
\begin{align}\label{eq:update-x}
    \begin{split}
    \hat{\vx}_i^{t+1} &= \hat{\vx}_i^{t} + Q\left(\vx_i^{t}-\hat{\vx}_i^{t}+\vxi_i^t\right)-\vxi_i^t,\\
    \vx_i^{t+1} &= \vx_i^{t} + \gamma\sum_{j=1}^{n} \mA_{ij}(\hat{\vx}_j^{t+1}-\hat{\vx}_i^{t+1}),
    \end{split}
\end{align}
\endgroup
where \mbox{$\hat{\vx}_i^0=\vect{0}$}, for all \mbox{$i\in[n]$}. Let \mbox{$\overline{\vx}^{t}$} be the average consensus at round $t$, i.e., \mbox{$\frac{1}{n}\sum_{i=1}^{n}\vx_i^{t}$}. Let matrix \mbox{$\mX^t\in\bbR^{n\times m}$} represents \mbox{$\left[\vx_1^t,\vx_2^t,\dots,\vx_n^t \right]^\top$}. Similarly, we denote $n\times m$ matrices \mbox{$\hat{\mX}^t=\left[\hat{\vx}_1^t,\hat{\vx}_2^t,\dots,\hat{\vx}_n^t\right]^\top$}, \mbox{$\overline{\mX}^t=\left[\overline{\vx}^t,\dots,\overline{\vx}^t\right]^\top$}, and \mbox{$\mZ^t=\left[\vxi_1^t,\vxi_2^t,\dots,\vxi_n^t \right]^\top$}. Thus, the update rule in Eq.~\eqref{eq:update-x} can be written in the following matrix-wise format:
\begin{align}\label{eq:update-x-matrix}
    \begin{split}
    \hat{\mX}^{t+1} &= \hat{\mX}^{t} + Q\left(\mX^{t}-\hat{\mX}^{t}+\mZ^t\right)-\mZ^t,\\
    \mX^{t+1} &= \mX^{t} + \gamma\left(\mA-\mI\right)\hat{\mX}^{t+1},
    \end{split}
\end{align}
where $Q(.)$ addresses rows of the matrix. Also, note that \mbox{$\overline{\mX}^t = \frac{1}{n}\vect{1}\vect{1}^\top\mX^t$} by definition. Hence, due to the fact that $\frac{1}{n}\vect{1}\vect{1}^\top \left(\mA-\mI\right) = \vect{0}$ and the consensus rule in Eq.~\eqref{eq:update-x-matrix}, we can conclude that \mbox{$\overline{\mX}^{t+1} = \overline{\mX}^{t}$}. Thus, a fixed average during the time, namely \mbox{$\overline{\vx}^t = \overline{\vx}$}. The following lemma describes the behaviour of the update rule in Eq.~\eqref{eq:update-x}.

\begin{lemma}[A variation of Theorem~2 from~\cite{koloskova2019decentralized}]\label{lem:modified-choco} The update rule in Eq.~\eqref{eq:update-x}, has the following property:
\begin{align*}
    e_t \leq \left(1-\frac{\delta^2\omega}{164}\right) e_{t-1} + L z_{t},
\end{align*}
when using the stepsize \mbox{$\gamma\triangleq\frac{\delta^2\omega}{32\delta+2\delta^2+8\beta^2+4\delta\beta^2-8\delta\omega}$}, where \mbox{$\beta = \left\lVert\mI-\mA\right\rVert_2$}, \mbox{$e_t \triangleq \bbE_{\vzeta}\left[\sum_{i=1}^{n}\left(\left\lVert \vx_i^{t} - \overline{\vx}\right\rVert^2 + \left\lVert \vx_i^{t} - \hat{\vx}_i^{t+1} \right\rVert^2\right)\right]$}, \mbox{$z_t \triangleq \sum_{i=1}^{n}\left\lVert \vxi_i^{t}\right\rVert^2$}, and \mbox{$L\triangleq{(1-\omega)(2-\omega)}/{\omega}$}.
\end{lemma}

The key difference between Lemma~\ref{lem:modified-choco} and~\cite[Theorem~2]{koloskova2019decentralized} is the existence of noise signals $\vxi_i^t$ in Eq.~\eqref{eq:update-x}, thus a slight difference in our proposed  upper bound. The proof for Lemma~\ref{lem:modified-choco} is provided in Appendix~\ref{app:modified-choco-proof}. We are now ready to present the proof for Theorem~\ref{thm:asymptotic}.

\begin{proof}(Theorem~\ref{thm:asymptotic})

For each hypothesis $\theta$, we divide Eq.~\eqref{eq:update-mu-1} and Eq.~\eqref{eq:update-mu-2} respectively by Eq.~\eqref{eq:update-nu-1} and Eq.~\eqref{eq:update-nu-2}, and take logarithm of the result, therefore

\begingroup
\allowdisplaybreaks
\begin{align}\label{eq:update-joint}
    \log\frac{\vmu_i^{t+1}}{\vnu_i^{t+1}} &= \log\frac{\vmu_i^{t}}{\vnu_i^{t}} + \gamma\sum_{j=1}^{n} \mA_{ij}\left(\log\frac{\hat{\vmu}_j^{t+1}}{\hat{\vnu}_j^{t+1}}-\log\frac{\hat{\vmu}_i^{t+1}}{\hat{\vnu}_i^{t+1}}\right), \nonumber\\
    \log\frac{\hat{\vmu}_i^{t+1}}{\hat{\vnu}_i^{t+1}} &= \log\frac{\hat{\vmu}_i^{t}}{\hat{\vnu}_i^{t}} + \log\frac{\hat{\vnu}_i^t}{\hat{\vnu}_i^{t+1}} \nonumber \\ &+ Q\left(\log\frac{\vmu_i^{t}}{\vnu_i^{t}}-\log\frac{\hat{\vmu}_i^{t}}{\hat{\vnu}_i^{t}}+\log\frac{\hat{\vnu}_i^{t+1}}{\hat{\vnu}_i^{t}}\right),
\end{align}
\endgroup
where all fractions are entrywise divisions as mentioned beforehand. Let's define \mbox{$x_i^{t}(\theta)=\log\left({\mu_i^{t}(\theta)}/{\nu_i^{t}(\theta)}\right)$} and \mbox{$\hat{x}_i^{t}(\theta)=\log\left({\hat{\mu}_i^{t}(\theta)}/{\hat{\nu}_i^{t}(\theta)}\right)$}, then Eq.~\eqref{eq:update-joint} turns into Eq.~\eqref{eq:update-x}. Hence, in line with Lemma~\ref{lem:modified-choco}, we know that \mbox{$\eta=1-({\delta^2\omega}/{164})$} so, \mbox{$\eta\in[0,1)$}. Lemma~\ref{lem:bounded-xi} also tells us that $z_t$ is bounded by $R^2$, then
\begin{align}\label{eq:e-t-bound}
    e_t \leq \eta^t e_0 + L\sum_{k=1}^{t} \eta^{t-k} z_k\leq\eta^t e_0 + \frac{LR^2(1-\eta^t)}{1-\eta},
\end{align}
where dividing by $t^2$, we can derive
\begingroup
\allowdisplaybreaks
\begin{align}
    \frac{1}{t^2}\bbE_{\vzeta}&\left[\left\lVert\left(\vx_i^{t} - \overline{\vx}\right)\right\rVert^2\right] \leq \frac{e_t}{t^2}\leq\frac{1}{t^2}\left(\eta^t e_0 + \frac{LR^2(1-\eta^t)}{1-\eta}\right) \Rightarrow\nonumber\\
    &
    \hspace{-1em}
    \lim_{t\to\infty}\frac{1}{t}\bbE_{\vzeta}\left[\left\lVert\left(\vx_i^{t} - \overline{\vx}\right)\right\rVert\right] = 0 \quad \mathrm{a.s.} \quad \text{for all } i\in[n],\label{eq:L2-x}
\end{align}
\endgroup
where $\overline{\vx} = \vect{0}$ by proper initialization, $\nu_i^0(\theta) = \mu_i^0(\theta)$, for all $i\in[n]$ and $\theta\in\Theta$. Equation~\eqref{eq:L2-x} suggests $L_2$ convergence where replacing $\vx_i^{t}$ with its equivalent expression (entrywise), we have \mbox{$\frac{1}{t}\bbE_{\vzeta}\left[\left\lVert\log\frac{\vmu_i^{t}(\theta)}{\vnu_i^{t}(\theta)}\right\rVert\right]\to 0$}. Also, due to the fact that $L_2$ convergence implies $L_1$ convergence~\cite[page 201]{resnick2019probability}, i.e., \mbox{$\frac{1}{t}\bbE_{\vzeta}\left[\left|\log\frac{\vmu_i^{t}(\theta)}{\vnu_i^{t}(\theta)}\right|\right]\to 0$}, it is inferred that for any $\epsilon>0$, there exists a $T$ such that for all $t\geq T$
\begin{align}\label{eq:ratio-1-2}
    \frac{1}{t}\left|\bbE_{\vzeta}\left[\log\frac{\mu_i^{t}(\theta)}{\nu_i^{t}(\theta)}\right]\right|\leq \frac{1}{t}\bbE_{\vzeta}\left[\left|\log\frac{\mu_i^{t}(\theta)}{\nu_i^{t}(\theta)}\right|\right]<\epsilon,
\end{align}
where the first inequality follows 
Jensen's inequality. We now consider the inequality in Eq.~\eqref{eq:ratio-1-2} for two arbitrary hypotheses $\theta_v\notin\Theta^\star$ and $\theta_w\in\Theta^\star$, therefore
\begingroup
\allowdisplaybreaks
\begin{align*}
    \frac{1}{t}\left|\bbE_{\vzeta}\left[\log\frac{\mu_i^{t}(\theta_v)}{\nu_i^{t}(\theta_v)} - \log\frac{\mu_i^{t}(\theta_w)}{\nu_i^{t}(\theta_w)}\right]\right| <& \\
    \frac{1}{t}\left|\bbE_{\vzeta}\left[ \log\frac{\mu_i^{t}(\theta_v)}{\nu_i^{t}(\theta_v)}\right]\right| + \frac{1}{t}\left|\bbE_{\vzeta}\left[\log\frac{\mu_i^{t}(\theta_w)}{\nu_i^{t}(\theta_w)}\right]\right| <& 2\epsilon,
\end{align*}
\endgroup
as a result of triangle inequality. Thus, we have:
\begin{align}\label{eq:diff-two-limits}
    \left|\frac{1}{t}\bbE_{\vzeta}\left[\log\frac{\mu_i^{t}(\theta_v)}{\mu_i^{t}(\theta_w)}\right] -  \frac{1}{t}\log\frac{\nu_i^{t}(\theta_v)}{\nu_i^{t}(\theta_w)}\right| < 2\epsilon,
\end{align}
so the two expressions inside Eq.~\eqref{eq:diff-two-limits}, converge to the same value in limit. Also, from Eq.~\eqref{eq:optimal-solution-ratio} we know that the second term converges to $-C_v$, almost surely, thus
\begin{align*}
    \lim_{t\to\infty}\frac{1}{t}\bbE_{\vzeta}\left[\log\frac{{\mu}_i^{t}(\theta_v)}{{\mu}_i^{t}(\theta_w)}\right]=-C_v,\quad \mathrm{a.s.},
\end{align*}
which is our desired result.
\end{proof}
The previous result shows the asymptotic exponential convergence of the ratio between any two non-optimal and optimal hypotheses almost surely. This means that in the limit, the set of optimal hypotheses dominates.

\section{Non-asymptotic Convergence Analysis}\label{sec:non-asymptotic-analysis}
In this section, we state a proof for Theorem~\ref{thm:non-asymptotic}, which provides an explicit convergence rate for our proposed algorithm in Eq.~\eqref{eq:update-mu-alg}. Lemma~\ref{lem:var-range} helps us to connect the convergence of the quantized process to that of the non-quantized process.

\begin{lemma}[Variation range]\label{lem:var-range}
Let Assumption~\ref{assump:init} holds. Further, assume that the update rules in Eq.~\eqref{eq:update-mu-alg} and Eq.~\eqref{eq:update-nu-alg} have the same initial values and the stepsize in Theorem~\ref{thm:non-asymptotic}. Then, for each agent $i$, parameter $\theta\in\Theta$, and $t=0,1,\dots$, the following inequality holds
\begin{align*}
    \left\lvert \bbE_{\vzeta}\left[\log\mu_i^t(\theta)\right]-\log\nu_i^t(\theta) \right\rvert \leq G_1 \quad \mathrm{const.}
\end{align*}
where $G_1=\frac{73\sqrt{nm}}{\delta^2\gamma\omega}\log\frac{1}{\alpha}$, with $\alpha=\min\{\alpha_1,\alpha_2\}$.
\end{lemma}

The proof for Lemma~\ref{lem:var-range} is provided in Appendix~\ref{app:var-range}. Now, we prove Theorem~\ref{thm:non-asymptotic}.

\begin{proof}(Theorem~\ref{thm:non-asymptotic})
By Lemma~\ref{lem:var-range}, for a non-optimal hypothesis $\theta_v\in\Theta^\star$ and an optimal hypothesis $\theta_w\in\Theta^\star$
\begingroup
\allowdisplaybreaks
\begin{align}
\left\lvert\bbE_{\vzeta}\left[\log\frac{\mu_i^t(\theta_v)}{\mu_i^t(\theta_w)}\right]-\log\frac{\nu_i^t(\theta_v)}{\nu_i^t(\theta_w)}\right\rvert &\leq \left\lvert\bbE_{\vzeta}\left[\log\frac{\mu_i^t(\theta_v)}{\nu_i^t(\theta_v)}\right]\right\rvert\label{eq:bounded-mu-nu-difference}\\
&
\hspace{-1em}
+\left\lvert\bbE_{\vzeta}\left[\log\frac{\mu_i^t(\theta_v)}{\nu_i^t(\theta_v)}\right]\right\rvert\leq 2G_1.\nonumber
\end{align}
\endgroup

Further, by~\cite[Lemma~10]{nedic2017fast}, the following inequality holds:
\begingroup
\allowdisplaybreaks
\begin{align*}
\bbE_{\theta_v,\theta_w}\left[ \log\frac{\nu_i^t(\theta_v)}{\nu_i^t(\theta_w)}\right] &\leq \max_{i}\log\frac{\nu_i^0(\theta_v)}{\nu_i^0(\theta_w)} + \frac{12\log n}{1-\lambda_2(\mB)}\log\frac{1}{\alpha_2}\\
&-t\min_{\substack{\theta_v\notin\Theta^\star}}\left(F(\theta_v)-F^\star\right),
\end{align*}
\endgroup
where the expectation $\bbE_{\theta_v,\theta_w}[.]$, indicates the randomness of observations. By Assumption~\ref{assump:init} and the definition of $\mB$, the above inequality can be modified as follows:
\begin{align}\label{eq:nu-ratio-expectation-upper-bound}
\bbE_{\theta_v,\theta_w}\left[ \log\frac{\nu_i^t(\theta_v)}{\nu_i^t(\theta_w)}\right] \leq -tC_1 + G_2,
\end{align}
where \mbox{$C_1=\min_{\substack{\theta_v\notin\Theta^\star}}\left(F(\theta_v)-F^\star\right)$} and \mbox{$G_2=\frac{16\log n}{\gamma\delta}\log\frac{1}{\alpha}$}. On the other hand, we have
\begin{align}\label{eq:log-tilde-mu-expectation-upperbound}
\bbE_{\vzeta}\left[\log\tilde{\mu}^t_i(\theta_v)\right]\leq \bbE_{\vzeta}\left[\log\frac{\tilde{\mu}^t_i(\theta_v)}{\tilde{\mu}^t_i(\theta_w)}\right]=\bbE_{\vzeta}\left[\log\frac{\mu^t_i(\theta_v)}{\mu^t_i(\theta_w)}\right],
\end{align}
thus the following inequalities hold:
\begingroup
\allowdisplaybreaks
\begin{align*}
&\bbP\left(\bbE_{\vzeta}\left[\log\tilde{\mu}_i^t(\theta_v)\right]\geq-\frac{t}{2}C_1+2G_1+G_2\right)\\
\leq&\bbP\left(\bbE_{\vzeta}\left[\log\frac{\mu^t_i(\theta_v)}{\mu^t_i(\theta_w)}\right]\geq-\frac{t}{2}C_1+2G_1+G_2\right)\\
\leq&\bbP\left(\log\frac{\nu_i^t(\theta_v)}{\nu_i^t(\theta_w)}\geq-\frac{t}{2}C_1+G_2\right)\\
+&\underbrace{\bbP\left(\bbE_{\vzeta}\left[\log\frac{\mu^t_i(\theta_v)}{\mu^t_i(\theta_w)}\right]\geq\log\frac{\nu_i^t(\theta_v)}{\nu_i^t(\theta_w)}+2G_1\right)}_{=0, \quad\text{by Eq.~\eqref{eq:bounded-mu-nu-difference}}}\\
\leq&\bbP\left(\log\frac{\nu_i^t(\theta_v)}{\nu_i^t(\theta_w)}-\bbE_{\theta_v,\theta_w}\left[\log\frac{\nu_i^t(\theta_v)}{\nu_i^t(\theta_w)}\right]\geq\frac{t}{2}C_1\right),
\end{align*}
\endgroup
where the first inequality follows Eq.~\eqref{eq:log-tilde-mu-expectation-upperbound}, the second inequality follows union bound, and the third inequality holds as a result of Eq.~\eqref{eq:nu-ratio-expectation-upper-bound}. With a similar approach to the proof for~\cite[Theorem~2]{nedic2017fast}, we can use McDiarmid's inequality to bound the above expression by probability $\rho$.  Therefore, for all \mbox{$t\geq T(\rho)= \frac{8}{C_1^2}\left(\log\alpha\right)^2\log\frac{1}{\rho}$}, with a probability at least $1-\rho$, the following inequality holds:
\begin{align*}
    \bbE_{\vzeta}\left[\log{\tilde{\mu}}_i^{t}(\theta_v)\right]\leq -\frac{t}{2}C_1 + 2G_1 + G_2 \leq -\frac{t}{2}C_1 + C_2,
\end{align*}
where $C_2=\frac{162\sqrt{nm}}{\delta^2\gamma\omega}\log\frac{1}{\alpha}$.
\end{proof}
We provided a probabilistic finite-time convergence result for our algorithm. The rate of convergence $C_1$ in our result is same as that of~\cite[Theorem~2]{nedic2017fast}, but the constant $C_2$ has worse dependencies on parameters $n,m,\delta$. However, we believe the bound is loose, and it can be greatly improved. This remains for future work.

\section{Numerical Experiments}
\label{sec:experiments}
Our analysis in Section~\ref{sec:non-asymptotic-analysis} suggests that with a high probability and after a sufficient time, the agents' beliefs linearly concentrate around the set of optimal hypotheses, under the update rule proposed in our algorithm (Eq.~\eqref{eq:update-mu-alg}). As illustrated in Theorem~\ref{thm:non-asymptotic}, the convergence rate depends on $C_1$ which is the difference between the first and second optimal values for function $F(\theta)$ in~\eqref{eq:non-bay-optim}, and $C_2$, which depends on the network's size and topology, number of beliefs, compression ratio, and initial conditioning of the problem. Furthermore, the validity of the bound is guaranteed after $T(\rho)$ rounds which also is inversely proportional to $C_1^2$. In this section, we quantify the performance through a series of empirical experiments. Specifically, we validate the compression capability of our algorithm for various network structures, the number of agents and beliefs, and several compression operator ratios.

We evaluate our algorithm on path and ring networks with low connectivity (i.e., $\delta^{-1}=\mcO(n^2)$), as well as Erd\H{o}s-R\'enyi (ER), torus, and complete topologies that have better connectivity. Topologies such as path and ring have mixing times of $O(n^2)$, which increases the message sharing dependency on the number of nodes quadratically. Therefore, the required number of iterations in such topologies is higher than their dense counterparts such as torus, fully connected (complete), or expander graphs~\cite{nedic2019graph}. The connectivity of Erd\H{o}s-R\'enyi graphs depends on their edge probabilities, thus we consider both $p_1=\mcO\left({\log n}/{n}\right)$ and $p_2=\mcO\left({1}/{\sqrt{n}}\right)$ in our experiments, where the former obviously provides sparser graphs. Note that from each of the two classes of ER graphs, we select a connected realization and fix it over the corresponding experiments. For a complete list of mixing times, see~\cite{nedic2019graph}. 

\textbf{Setup:} We begin by considering a group of $n=100$ agents respectively on path, ring, ER-$p_1$, torus, ER-$p_2$, and complete topologies with weights ${1}/{\max\{d_i+1,d_j+1\}}$ for off-diagonal pairs $(i,j)\in\mcE$, where $d_i$ is the degree of agent $i$. We select a large hypothesis set with $m=400$ as well as a finite set of observations $\mcS_i$ with size $|\mcS_i|=20$ for each agent $i$. We also generate a family of random probability distributions, $\{\ell_i\left(.|\theta_k\right)\}_{k=1}^m$, as well as a random $f_i$ on each $\mcS_i$ such that $|\Theta^\star|=1$. This way, the optimal hypothesis is not required to be locally optimal for all agents, by construction.

We partition our numerical simulations into two components. First, we fix a sequence of observations, sampled from $f_i$, for each agent $i\in[n]$, and apply our algorithm with one of the compression operators. In the case of randomized operators such as $\mathrm{qsgd}_{k\,\mathrm{bits}}$, and $\mathrm{rand}_{100\omega\%}$, we repeat the experiment $10$ times and compute the average as the final performance. This would show how randomness in the operators affects the behavior of the individual belief realization.  Second, we run $100$ Monte Carlo runs of the algorithm with a real-time sampling of the observations.

\begin{figure}[!t]
    \begin{minipage}{\linewidth}
    \begin{minipage}{0.44\linewidth}
    \includegraphics[width=\textwidth]{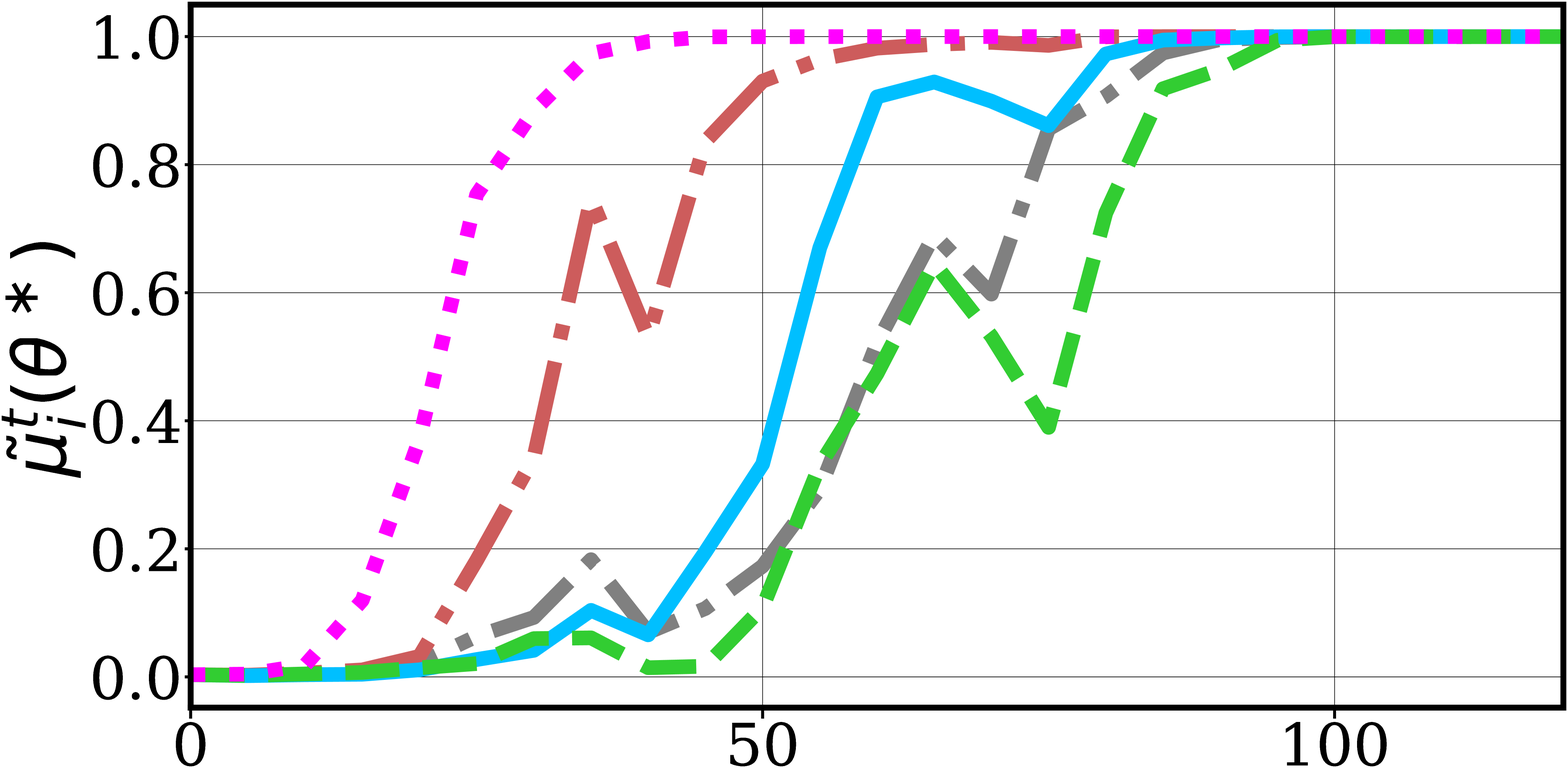}
    \end{minipage}
    \begin{minipage}{0.44\linewidth}
    \includegraphics[width=\textwidth]{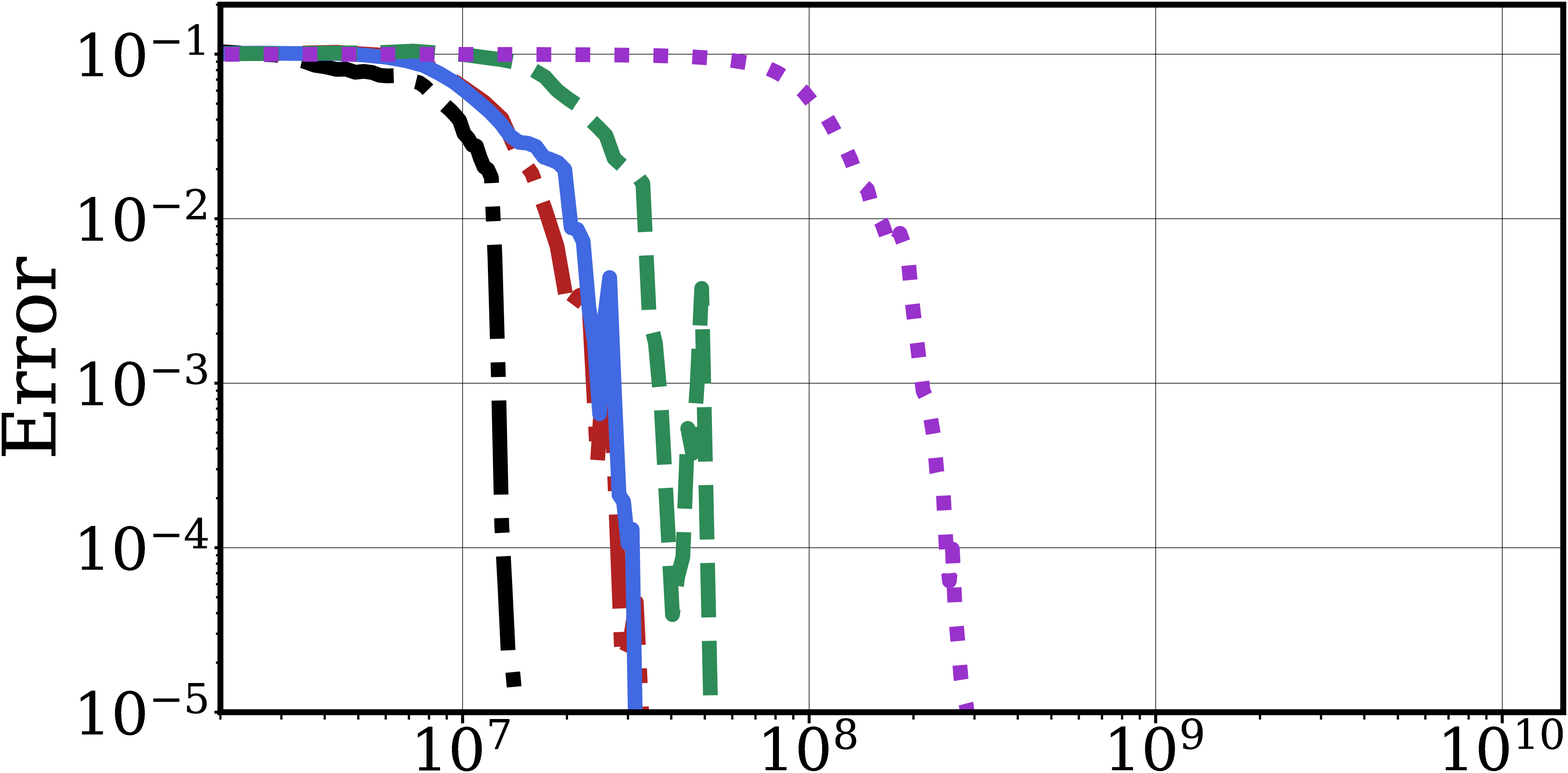}
    \end{minipage}
    \rotatebox{-90}{
    \hspace{-2.85em}\begin{minipage}{0.15\linewidth}
    \subcaption{\footnotesize Path}
    \label{fig:belief-path-100}
    \end{minipage}
    }
    \end{minipage}

    \begin{minipage}{\linewidth}
    \begin{minipage}{0.44\linewidth}
    \includegraphics[width=\textwidth]{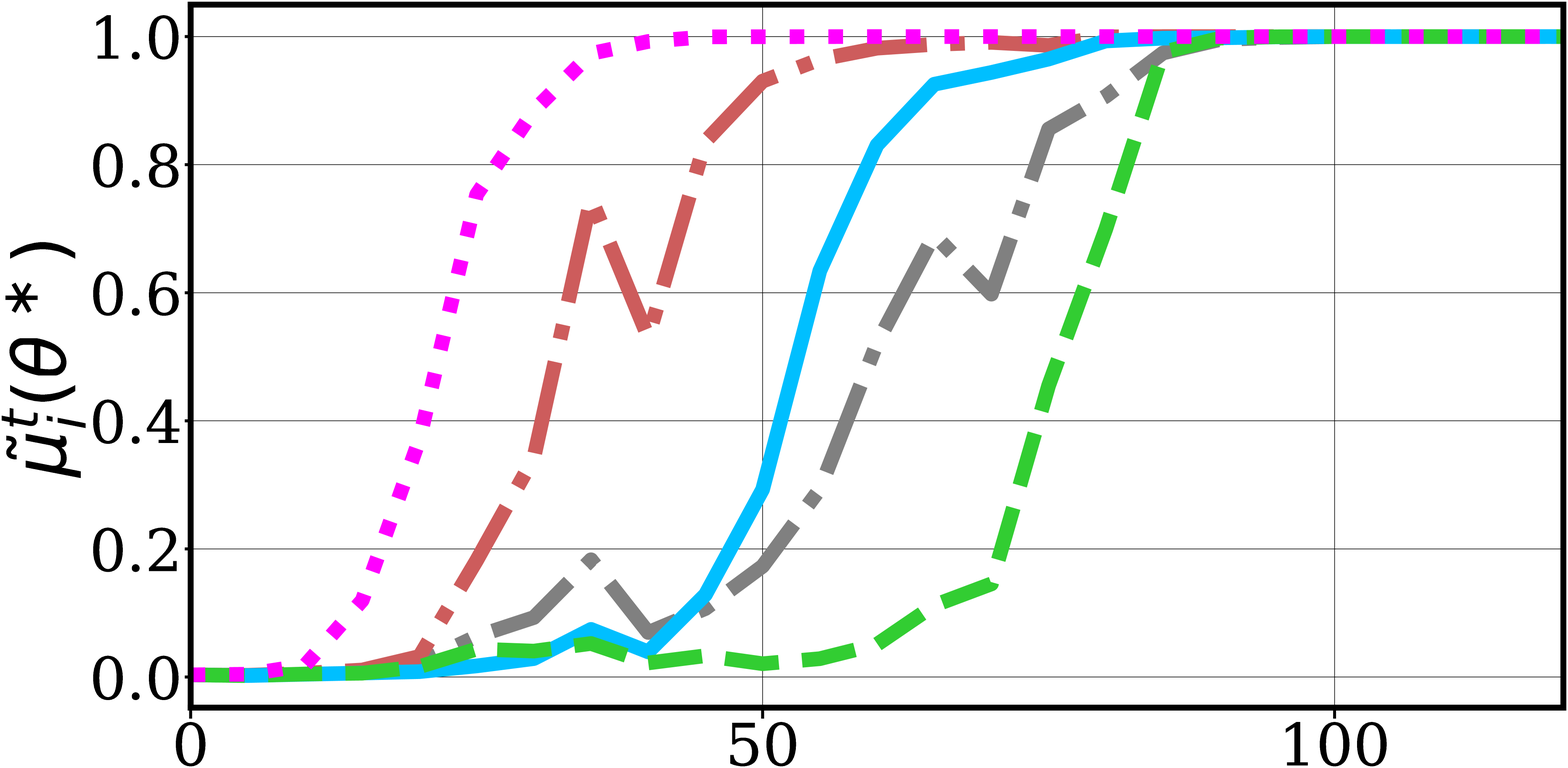}
    \end{minipage}
    \begin{minipage}{0.44\linewidth}
    \includegraphics[width=\textwidth]{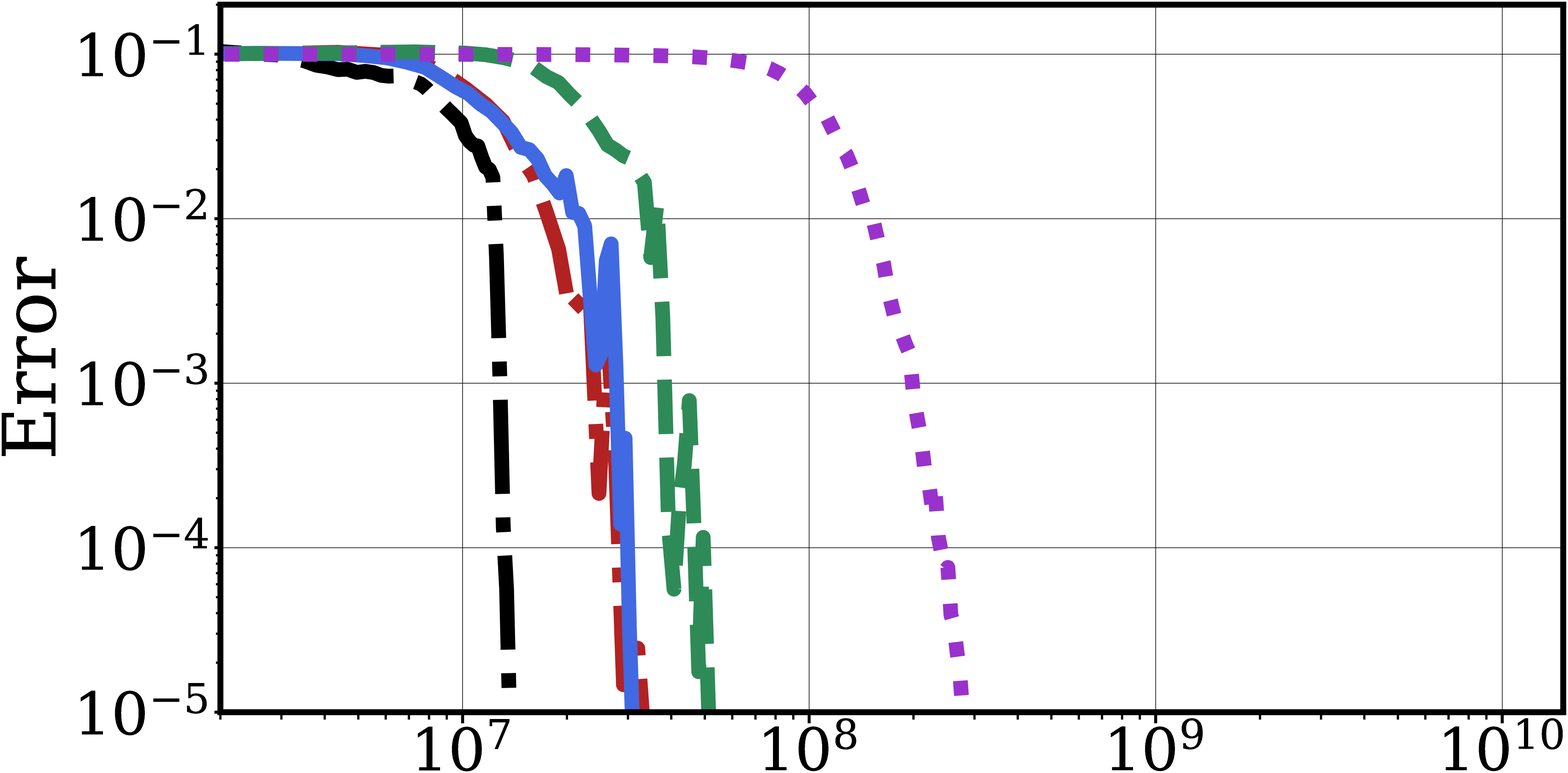}
    \end{minipage}
    \rotatebox{-90}{
    \hspace{-2.75em}\begin{minipage}{0.15\linewidth}
    \subcaption{\footnotesize Ring}
    \label{fig:belief-ring-100}
    \end{minipage}
    }
    \end{minipage}

    \begin{minipage}{\linewidth}
    \begin{minipage}{0.44\linewidth}
    \includegraphics[width=\textwidth]{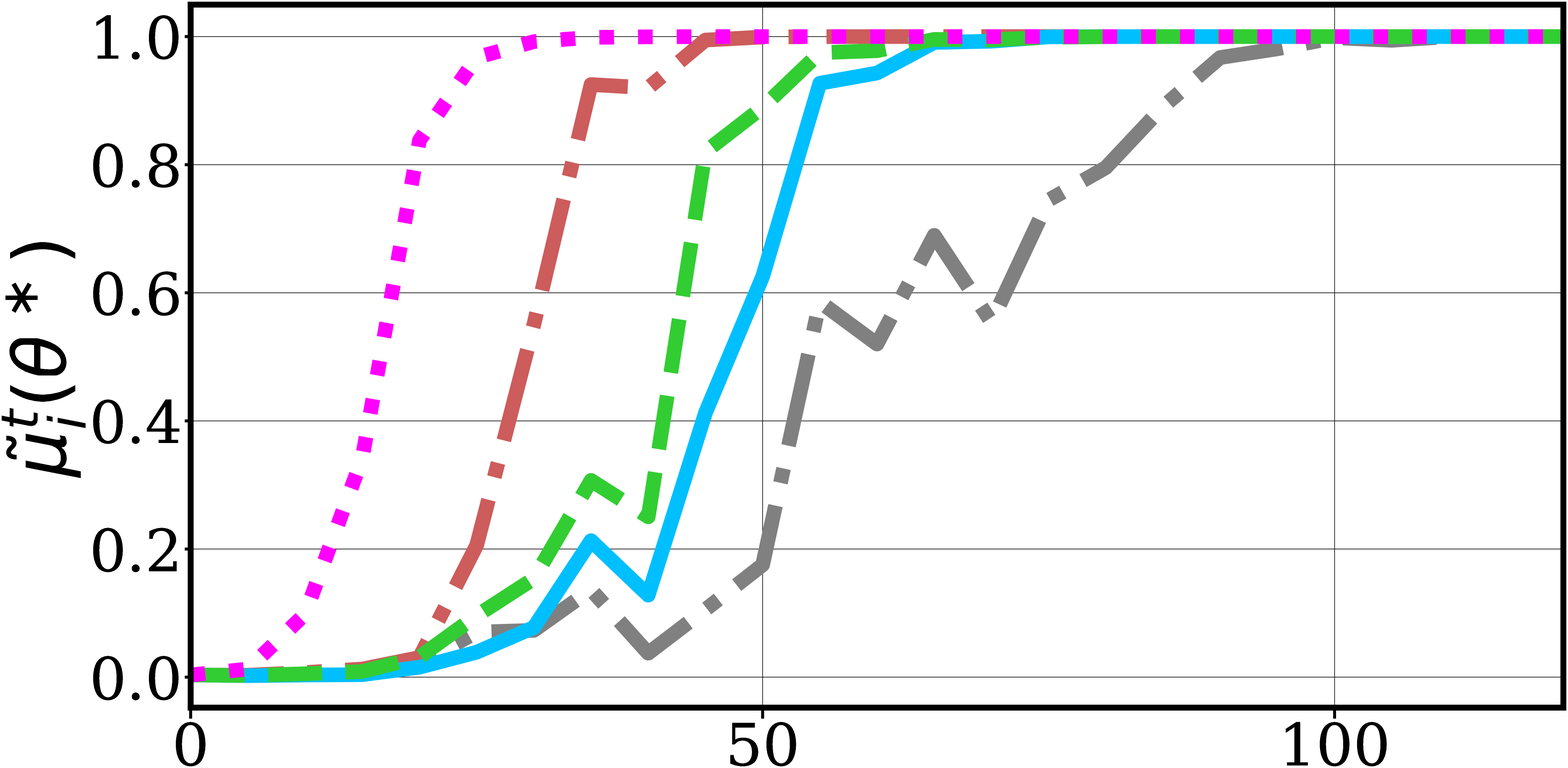}
    \end{minipage}
    \begin{minipage}{0.44\linewidth}
    \includegraphics[width=\textwidth]{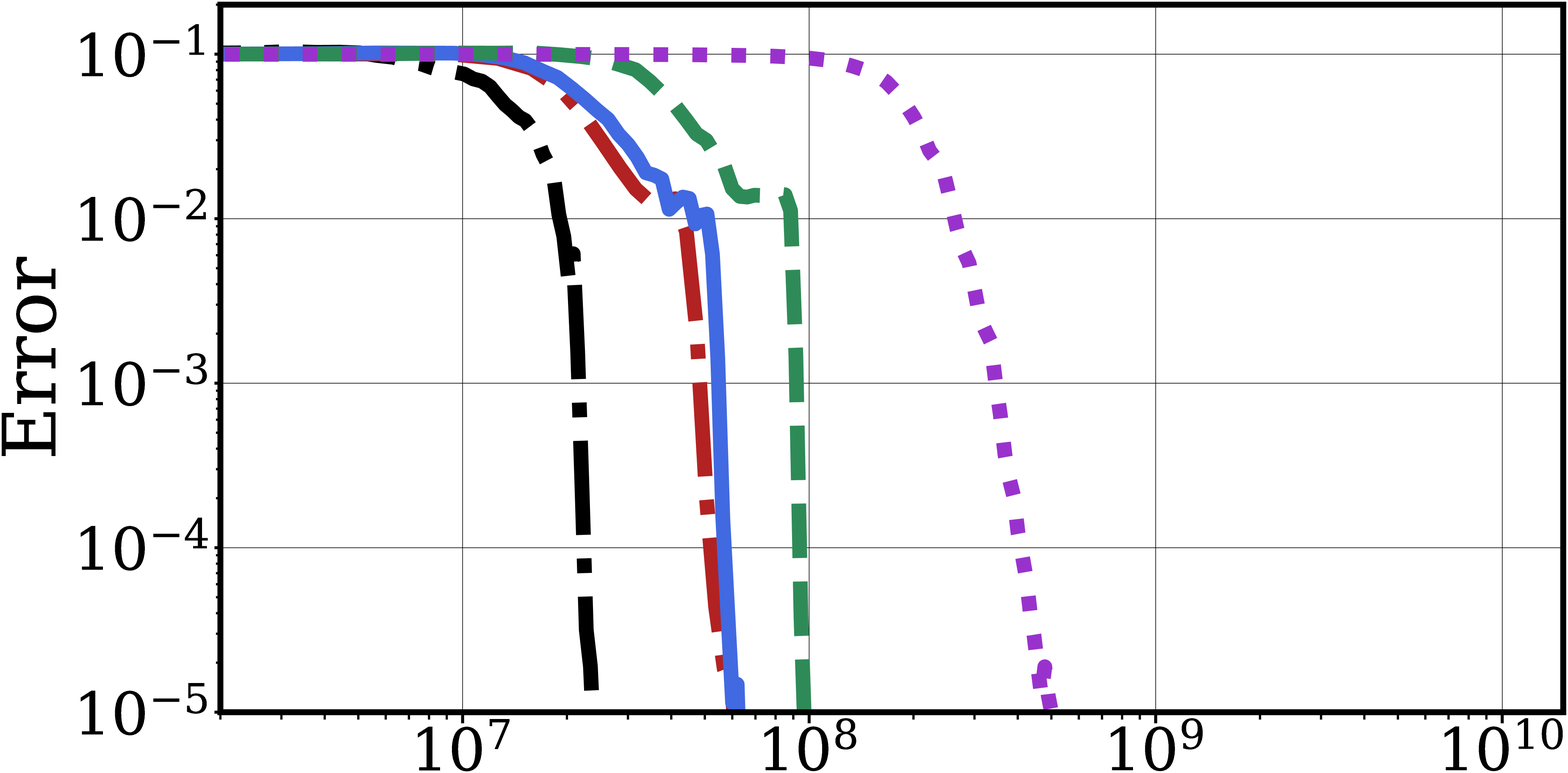}
    \end{minipage}
    \rotatebox{-90}{
    \hspace{-3.4em}\begin{minipage}{0.2\linewidth}
    \subcaption{{\footnotesize ER-$p_1$}}
    \label{fig:belief-erdos-renyi-logn-n-100}
    \end{minipage}
    }
    \end{minipage}
    
    \hspace{-0.35em}
    \begin{minipage}{\linewidth}
    \begin{minipage}{0.44\linewidth}
    \includegraphics[width=\textwidth]{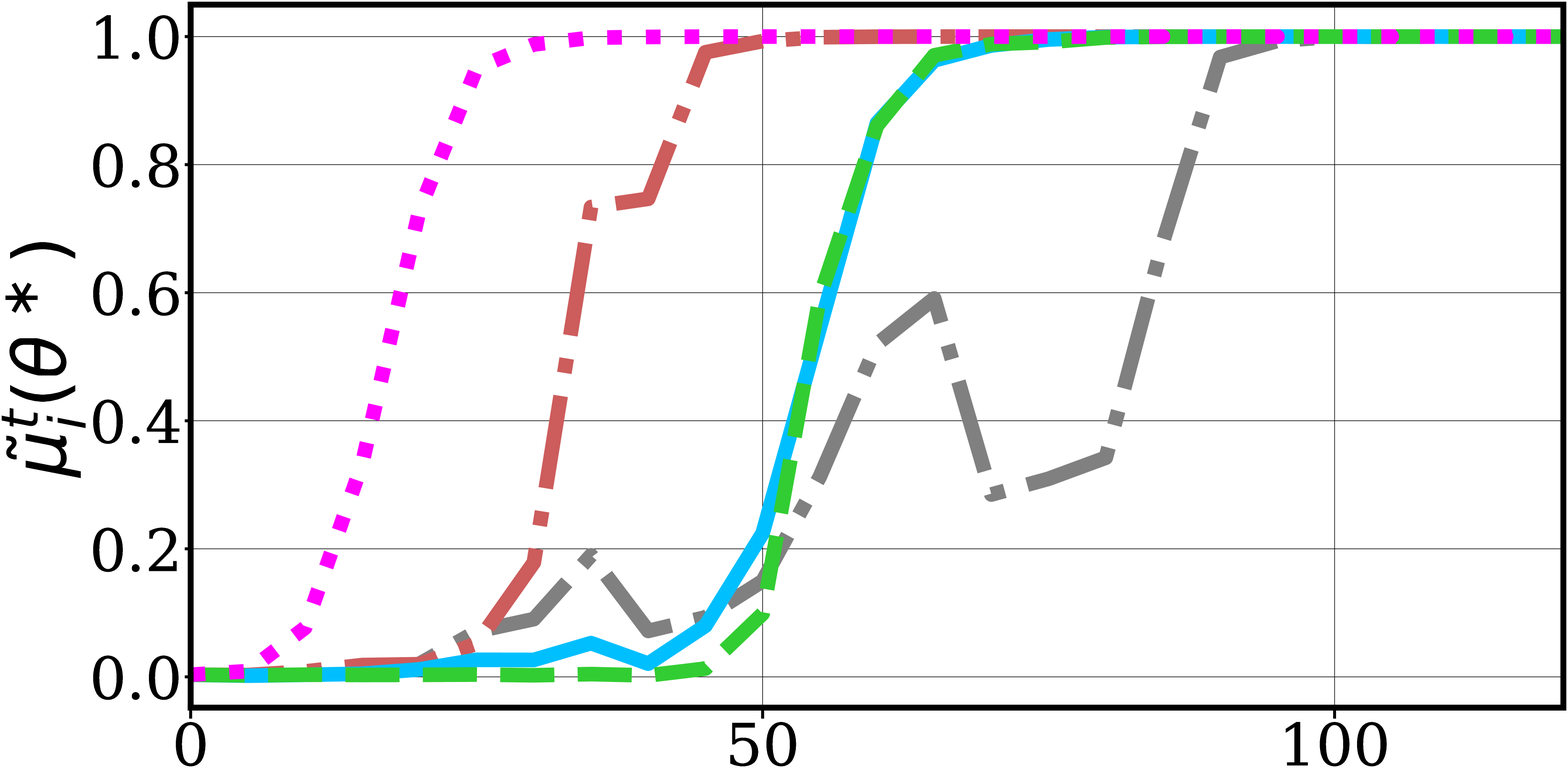}
    \end{minipage}
    \begin{minipage}{0.44\linewidth}
    \includegraphics[width=\textwidth]{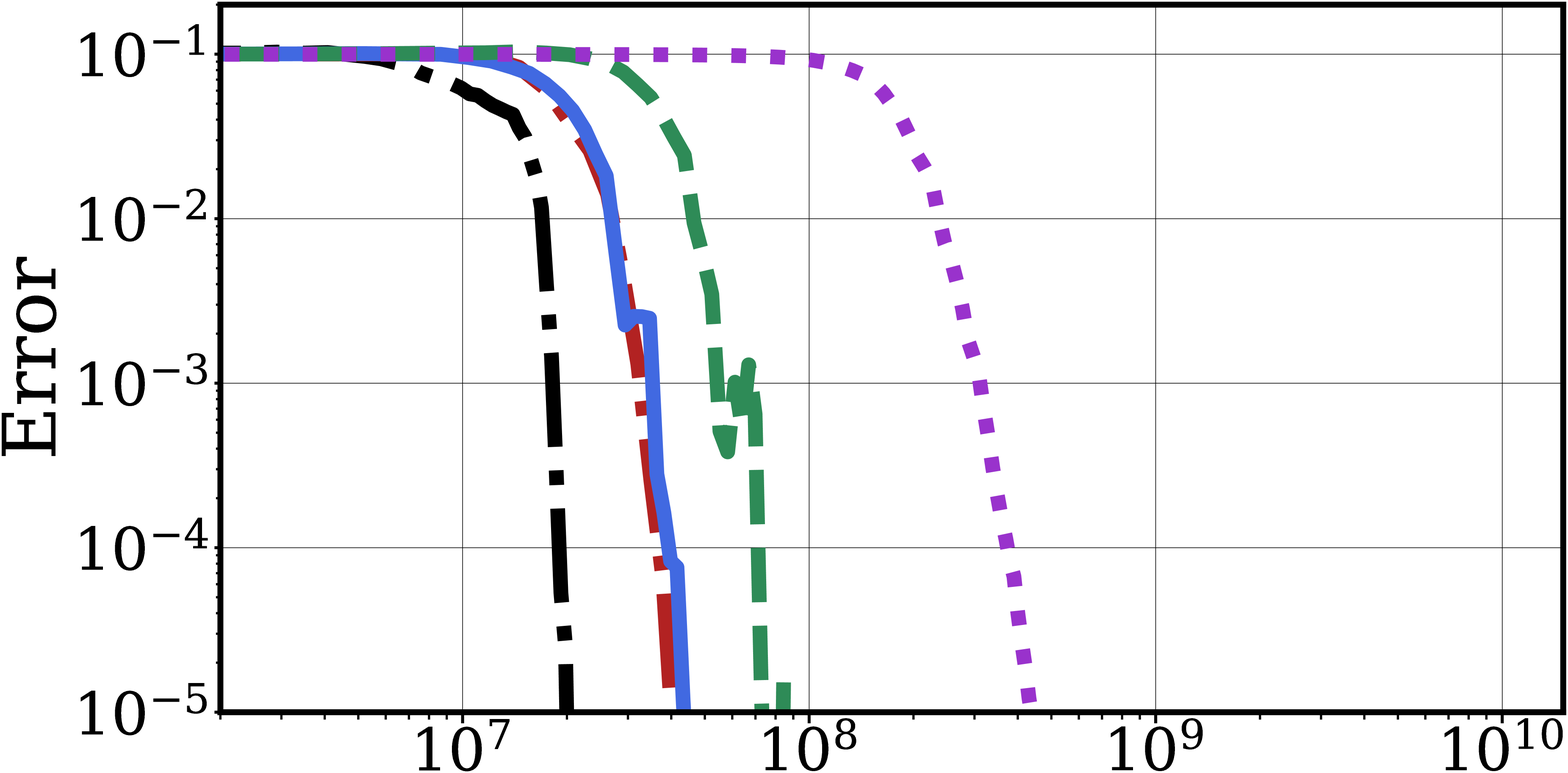}
    \end{minipage}
    \rotatebox{-90}{
    \hspace{-3.2em}\begin{minipage}{0.2\linewidth}
    \subcaption{\footnotesize Torus}
    \label{fig:belief-torus-100}
    \end{minipage}
    }
    \end{minipage}
    
    \begin{minipage}{\linewidth}
    \begin{minipage}{0.44\linewidth}
    \includegraphics[width=\textwidth]{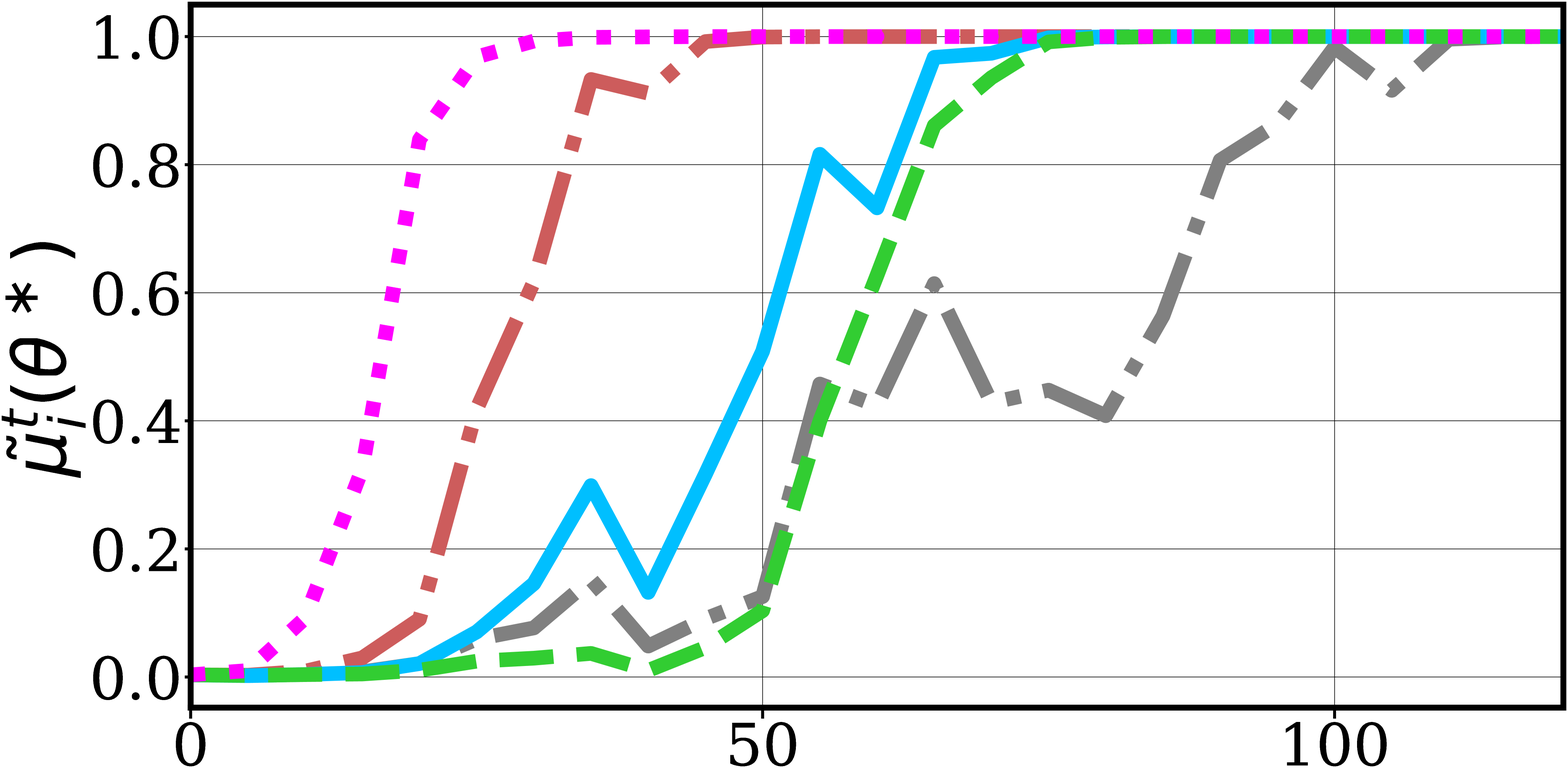}
    \end{minipage}
    \begin{minipage}{0.44\linewidth}
    \includegraphics[width=\textwidth]{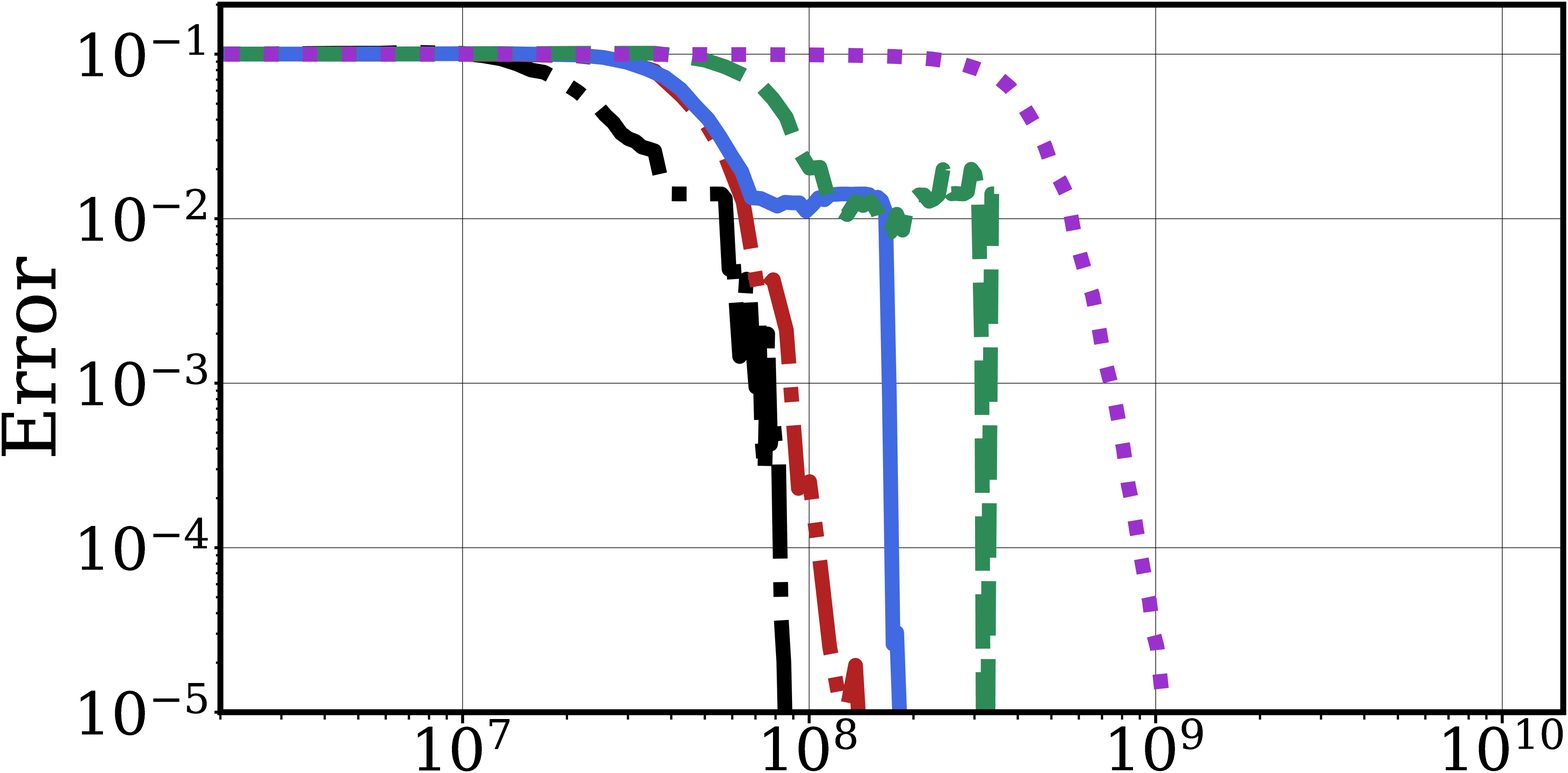}
    \end{minipage}
    \rotatebox{-90}{
    \hspace{-3.2em}\begin{minipage}{0.2\linewidth}
    \subcaption{{\footnotesize ER-$p_2$}}
    \label{fig:belief-erdos-renyi-sqrtn-n-100}
    \end{minipage}
    }
    \end{minipage}
    
    \hspace{-0.35em}
    \begin{minipage}{\linewidth}
    \begin{minipage}{0.44\linewidth}
    \includegraphics[width=\textwidth]{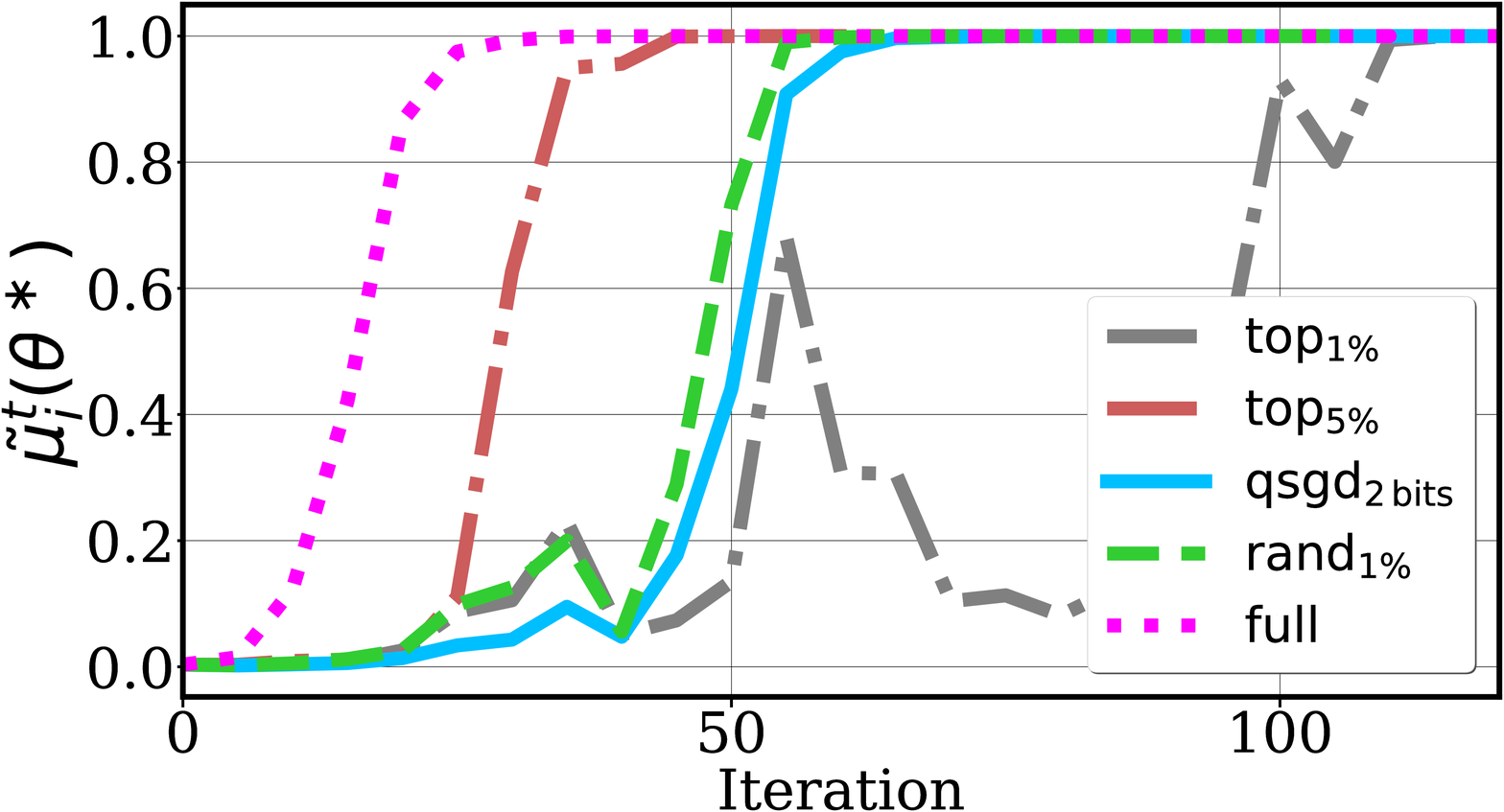}
    \end{minipage}
    \begin{minipage}{0.44\linewidth}
    \includegraphics[width=\textwidth]{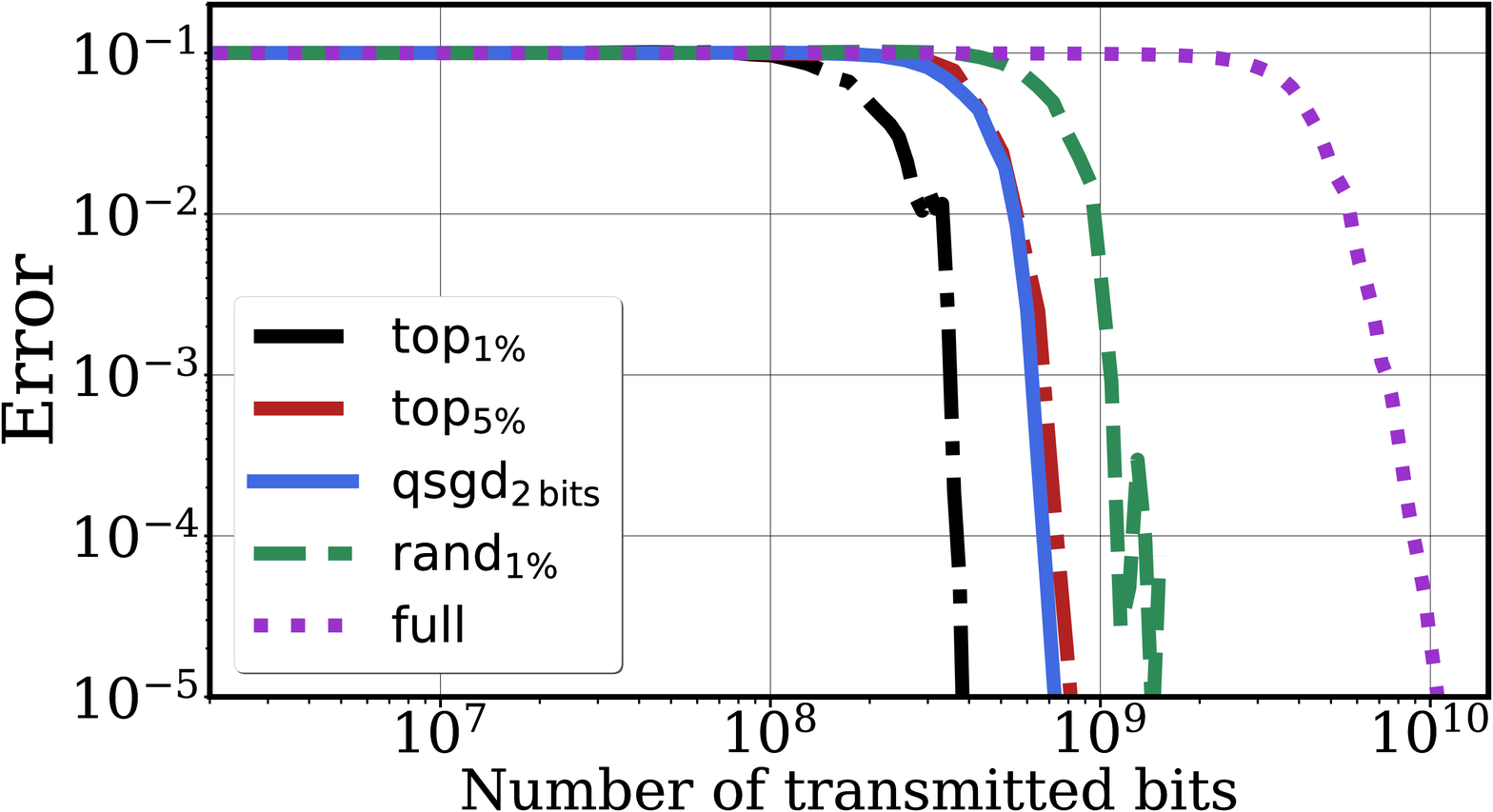}
    \end{minipage}
    \rotatebox{-90}{
    \hspace{-3.5em}\begin{minipage}{0.2\linewidth}
    \subcaption{\footnotesize Complete}
    \label{fig:belief-complete-100}
    \end{minipage}
    }
    \end{minipage}
    
    \caption{Plots corresponding to the randomized operators are the average of $10$ simulations. Convergence of the update rule in Eq.~\eqref{eq:update-mu-alg} using compression operators in Table~\ref{tab:compression} with the update rule in~\cite[Eq.~(2)]{nedic2017fast} over (a) path, (b) ring, (c) Erd\H{o}s-R\'enyi with edge probability \mbox{$p_1=\mcO\left({\log n}/{n}\right)$}, (d) torus, (e) Erd\H{o}s-R\'enyi with edge probability \mbox{$p_2=\mcO\left({1}/{\sqrt{n}}\right)$}, and (f) complete topologies, with $n=100$ agents and $m=400$ hypotheses. In all experiments, we consider a fixed sequence of observations for the agents, and select $\gamma$ by a grid search. (\textbf{left}) Belief evolution of one agent on the optimal hypothesis. (\textbf{right}) Convergence error per the number of transmitted bits through the corresponding network.}
    \label{fig:operators-comparison}

\end{figure}

\textbf{Beliefs Evolution:} We first compare the performance of our algorithm using the three compression operators introduced in Table~\ref{tab:compression} versus the update rule with perfect communication~\cite[Eq.~(2)]{nedic2017fast}. For this experiment, we consider the scenario with a fixed path of observations for the agents. We consider $\mathrm{qsgd}_{k\,\mathrm{bits}}$ with $k=2$ bits (least precision) and $b=64$ bits baseline. According to Eq.~\eqref{eq:qsgd}, this quantization satisfies Eq.~\eqref{eq:q-comp} with $\omega\approx 0.05$, so we select the other two operators $\mathrm{rand}_{5\%}$ and $\mathrm{top}_{5\%}$ with the same omega. We also consider $\mathrm{top}_{1\%}$ with a lower compression ratio. For the deterministic operators, we run the corresponding algorithm only once, but for the randomized operators with repeat the simulations $10$ times and consider the average. For the choice of $\gamma$ in all experiments, we apply a grid line search over the set \mbox{$\{{\omega}/{4},{\omega}/{2},\omega,2\omega,4\omega\}$} to pick the value that converges faster in practice. Figure~\ref{fig:operators-comparison} illustrates the result of this experiment, wherefrom the top (path graph) to the bottom (complete graph), the connectivity among agents increases, i.e., the spectral gap grows.

In the left-hand plots of Fig.~\ref{fig:operators-comparison}, we see the belief evolution of one agent on the optimal hypothesis for different compression operators. In the right-hand plots, we report the convergence error \mbox{$\frac{1}{n}\sum_{i=1}^{n}\left\lVert\tilde{\vmu}^t_i-\vmu^\star\right\rVert$}, per the number of transmitted bits, where $\vmu^\star$ is the vector with $1$ at the entry corresponding to $\theta^\star$ and the rest zero. As shown for example in the torus graph results (Fig.~\ref{fig:belief-torus-100}, our proposed algorithm requires less communication cost (about $5\%-20\%$ of the $\mathrm{full}$) to reach an $\epsilon=10^{-5}$ error compared to the algorithm in~\cite[Eq.~(2)]{nedic2017fast}. Nonetheless, the number of iterations required for convergence grows which is an inevitable consequence of arbitrary compression. Further, we can see that $\mathrm{qsgd}_{2\mathrm{bits}}$ converges faster than $\mathrm{top}_{5\%}$, but with decreasing the compression ratio $\omega$ to $0.01$, the $\mathrm{top}_{1\%}$ achieves a better performance than $\mathrm{qsgd}_{2\mathrm{bits}}$ in terms of the transmitted bits, but requires more iterations to converge (left figures). Complied to our intuition, $\mathrm{qsgd}_{2\mathrm{bits}}$, $\mathrm{top}_{5\%}$, and $\mathrm{top}_{1\%}$  outperform $\mathrm{rand}_{5\%}$ operator.

\begin{figure}[!ht]
    \raggedleft
    \begin{minipage}{0.33\linewidth}
    \raggedleft
    \includegraphics[width=\linewidth]{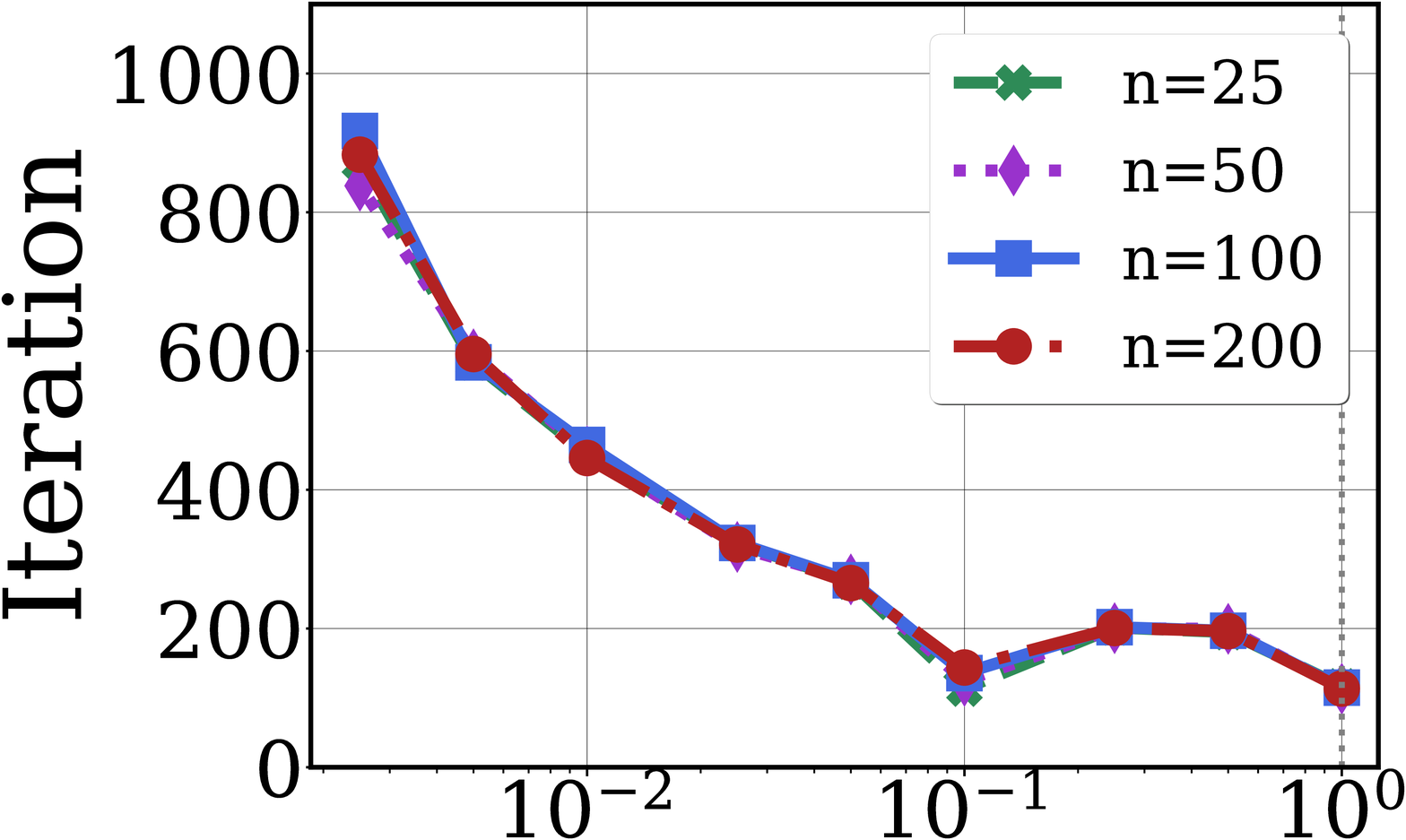}
    \includegraphics[width=\linewidth]{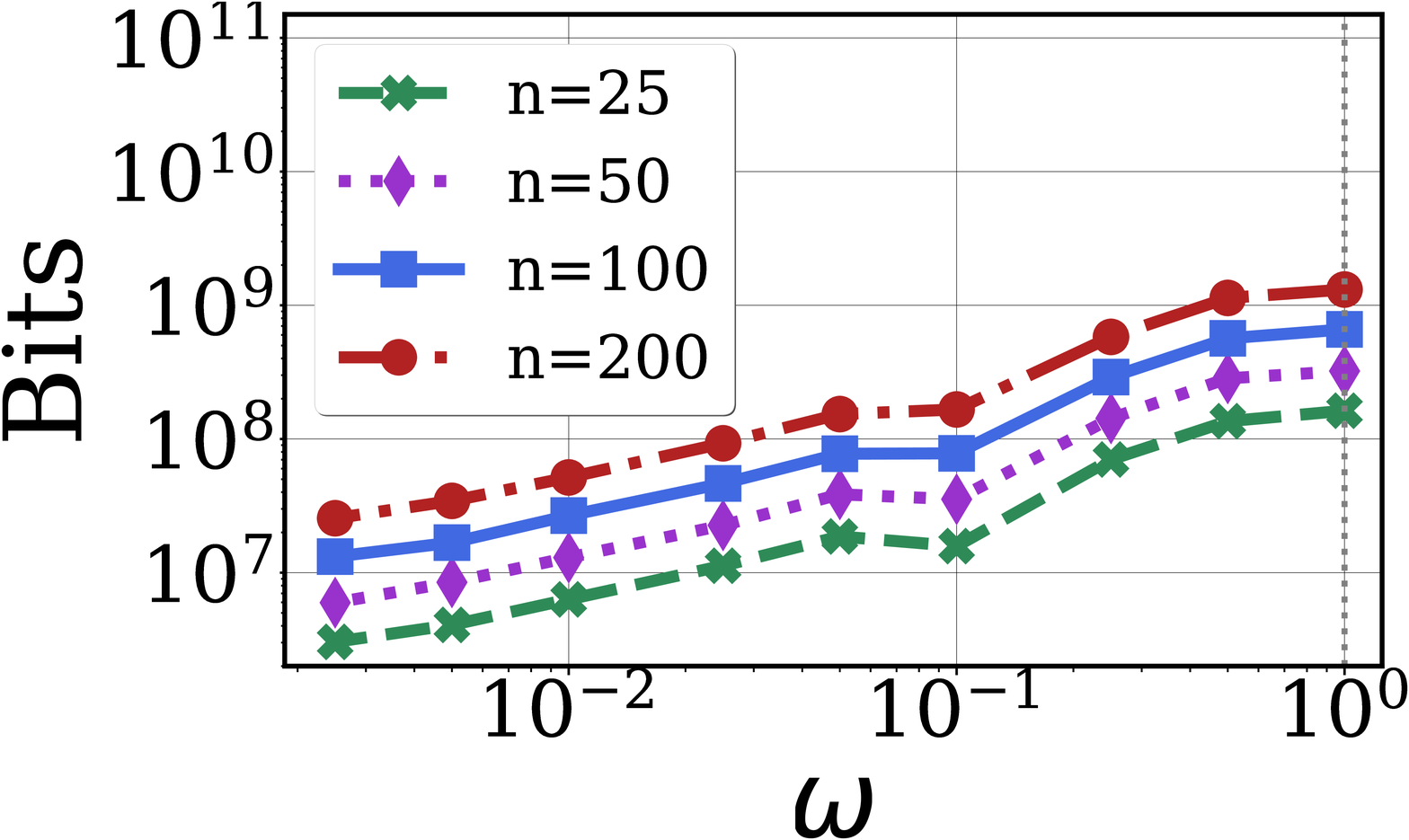}
    \subcaption{Path}
    \label{fig:n-top-omega-path}
    \end{minipage}
    \hspace{0.2em}
    \begin{minipage}{0.3\linewidth}
    \raggedleft
    \includegraphics[width=\linewidth]{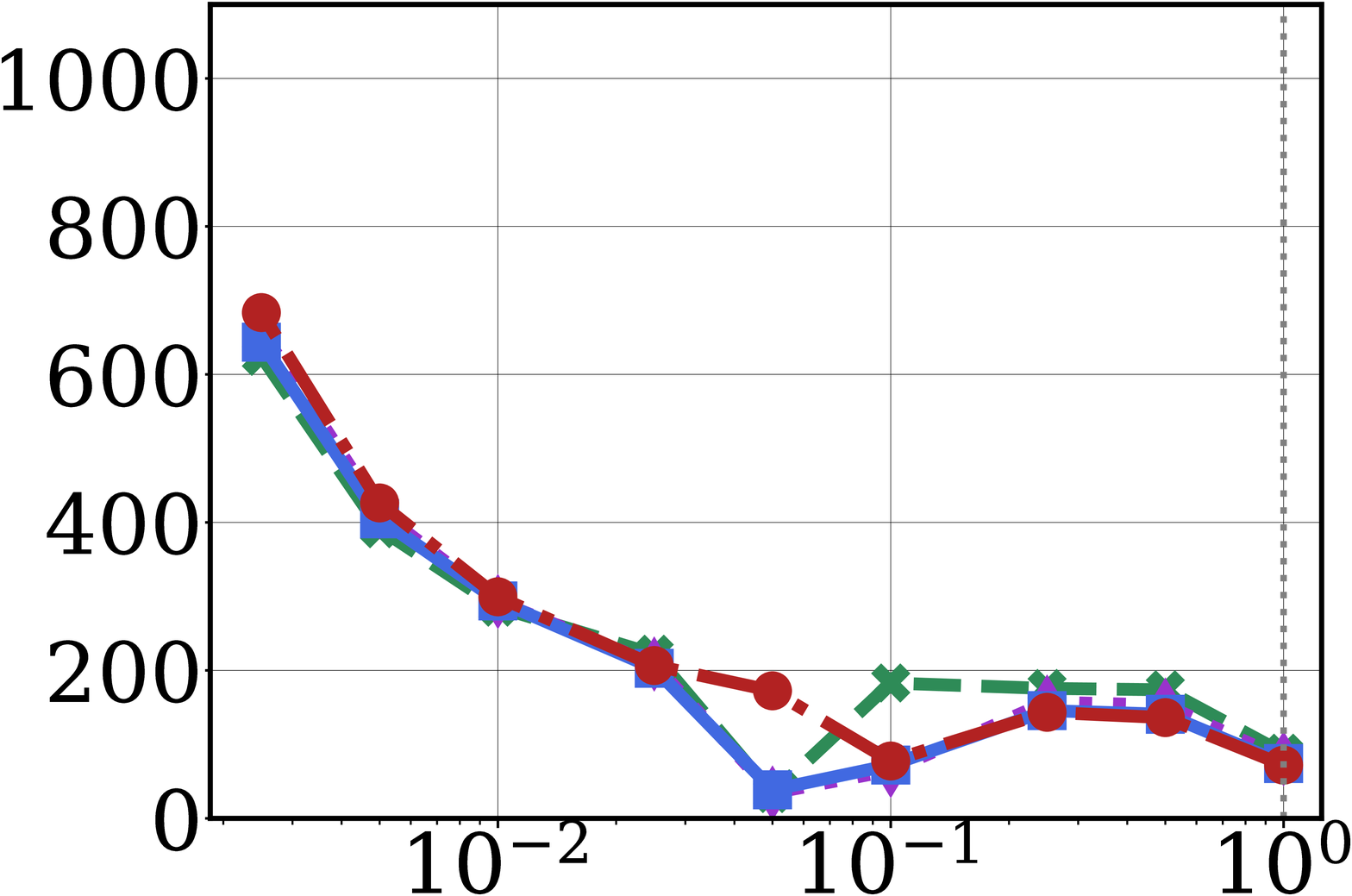}
    \includegraphics[width=\linewidth]{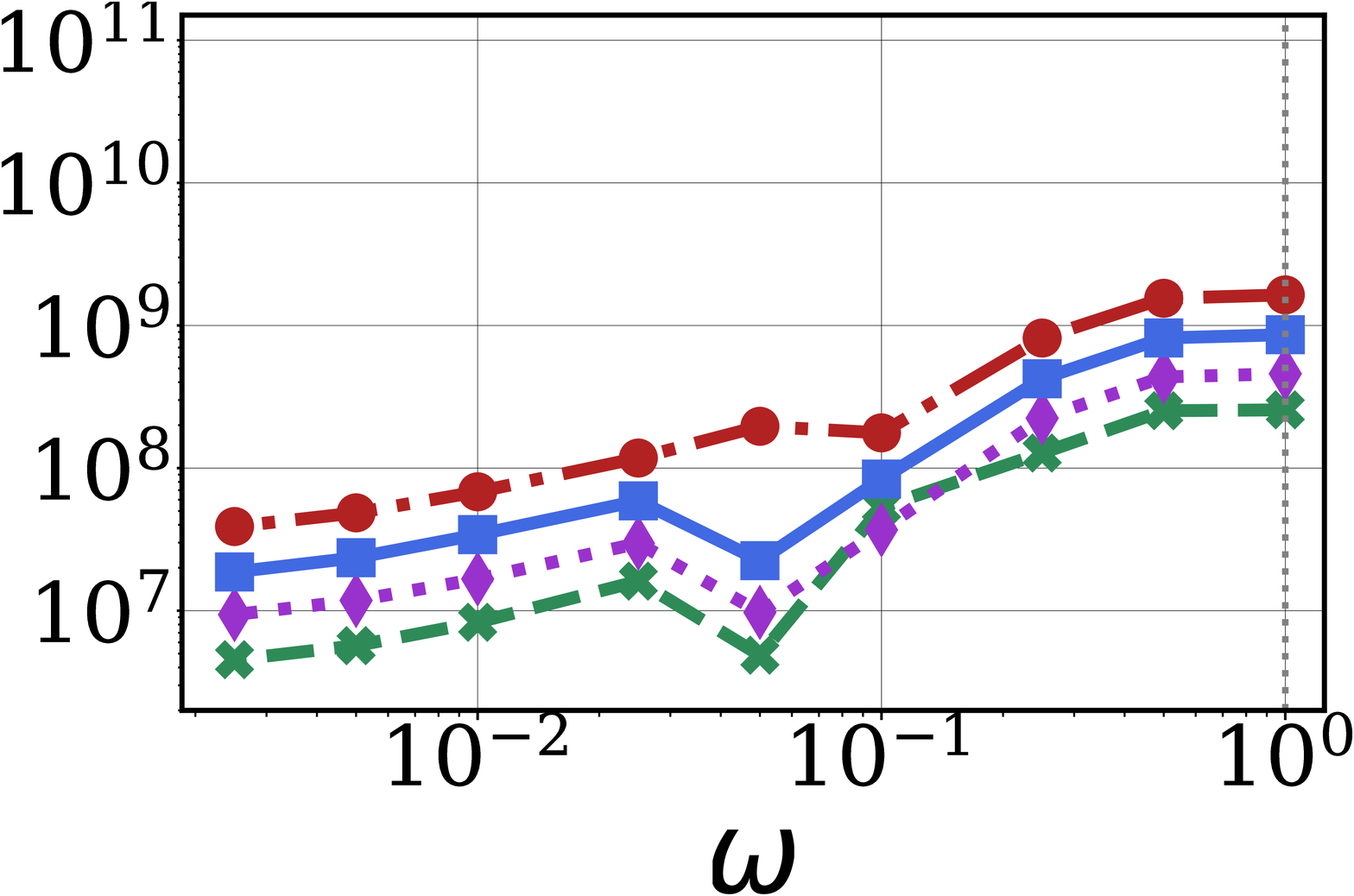}
    \subcaption{Torus}
    \label{fig:n-top-omega-torus}
    \end{minipage}
    \hspace{0.2em}
    \begin{minipage}{0.3\linewidth}
    \raggedleft
    \includegraphics[width=\linewidth]{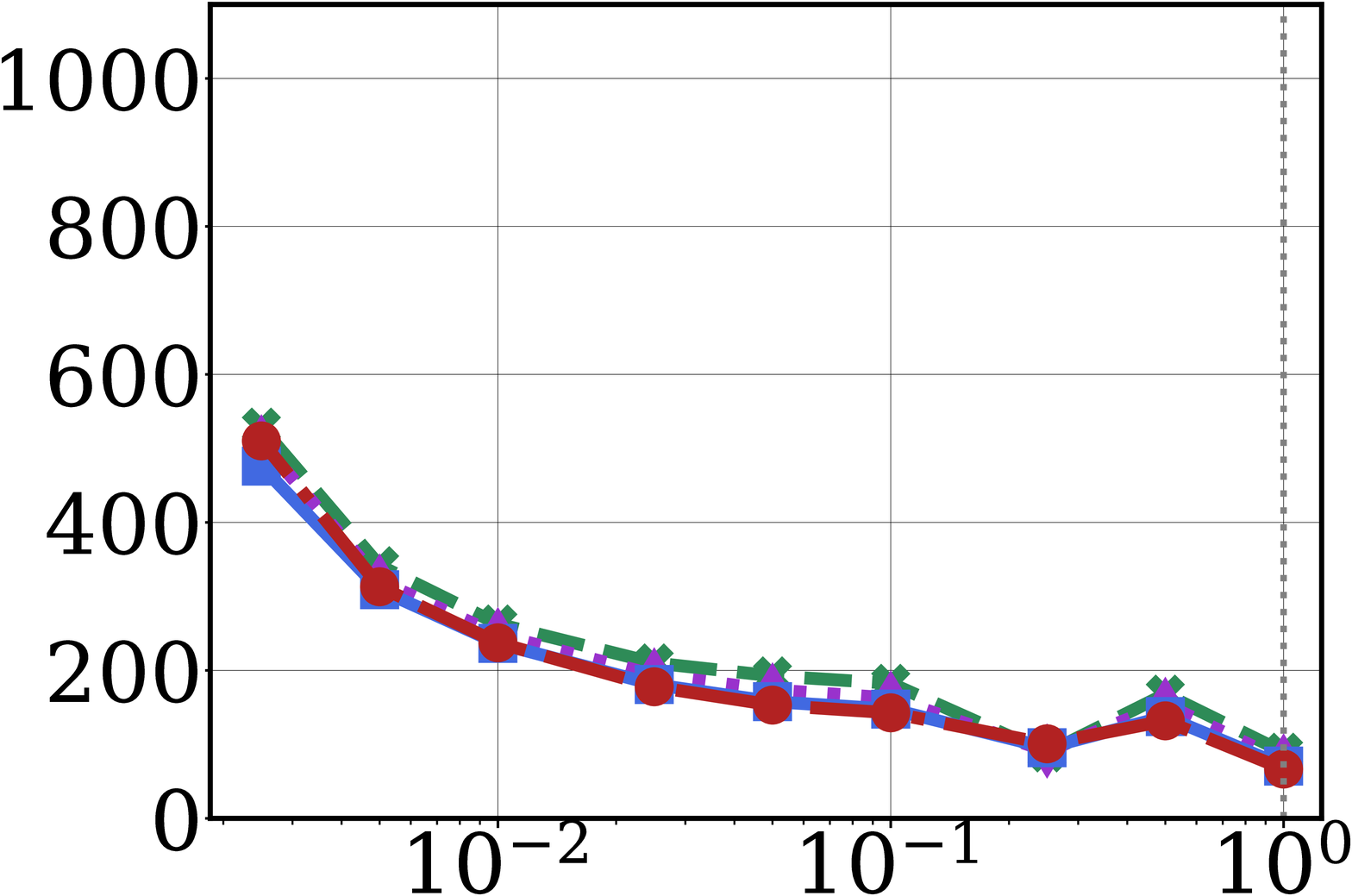}
    \includegraphics[width=\linewidth]{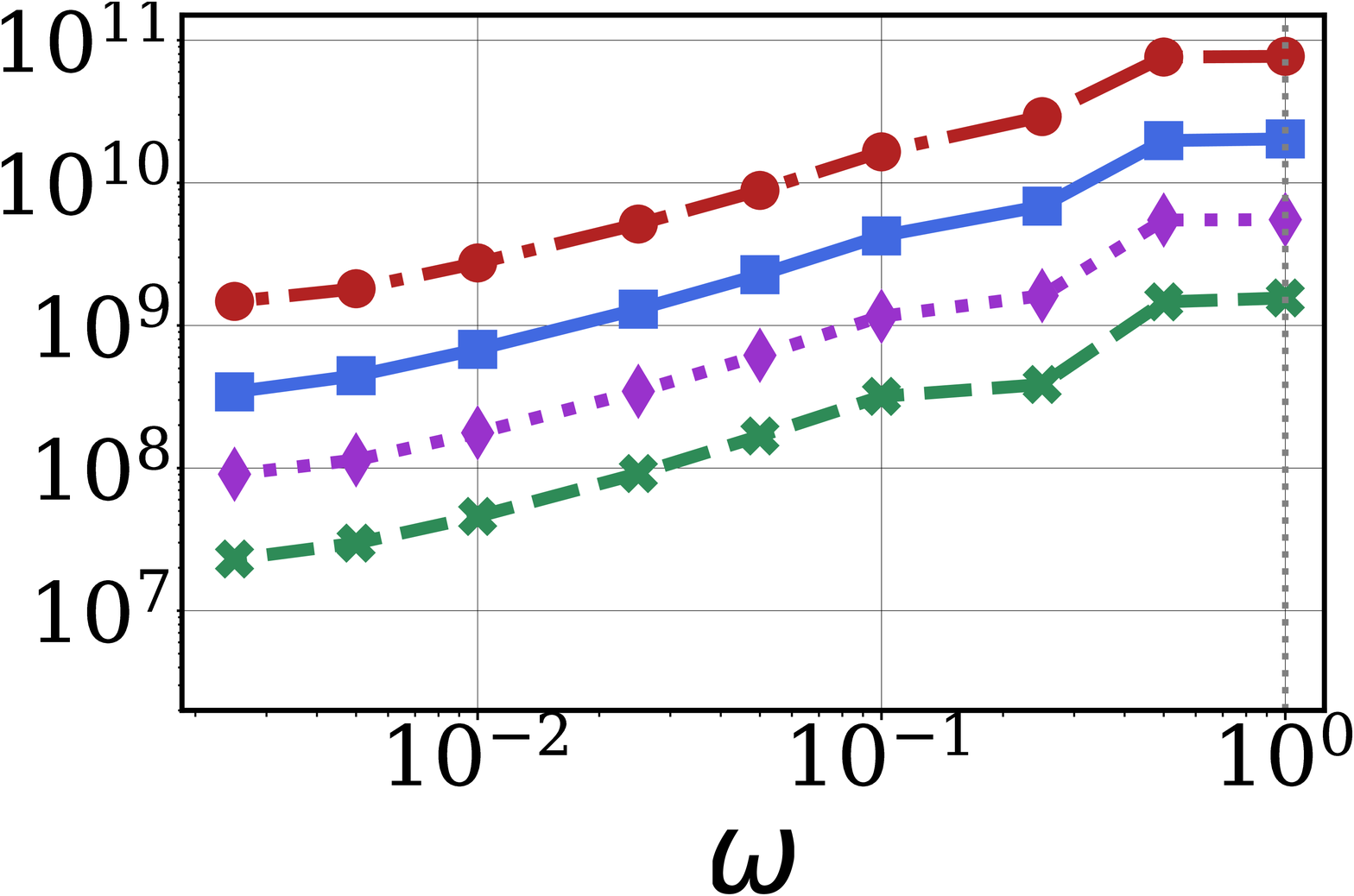}
    \subcaption{Complete}
    \label{fig:n-top-omega-complete}
    \end{minipage}
    \caption{The empirical mean over $100$ Monte Carlo runs of the number of iterations (top) and transmitted bits (bottom), required for \mbox{$\tilde{\mu}_i^t(\theta)<\epsilon$} for all agents on every \mbox{$\theta\notin\Theta^\star$} with $m=400$ hypotheses and \mbox{$\mathrm{top}_{100\omega\%}$} operator over (a) path, (b) torus, and (c) complete topologies. Each line corresponds to a separate value of $n$ for different values of $\omega$.}
    \label{fig:n-top}
\end{figure}

\begin{figure}
    \raggedleft
    \begin{minipage}{0.33\linewidth}
    \raggedleft
    \includegraphics[width=\linewidth]{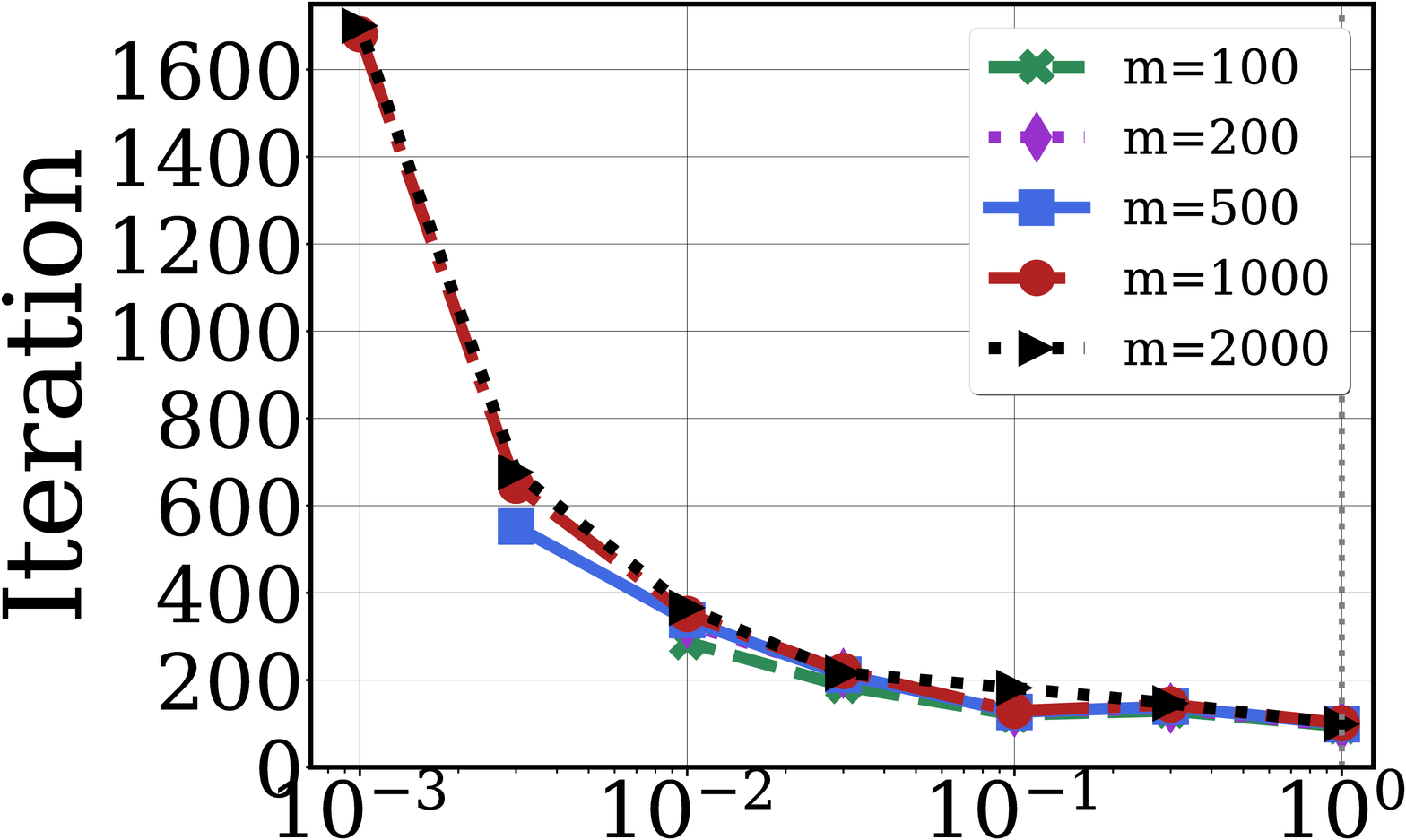}
    \includegraphics[width=\linewidth]{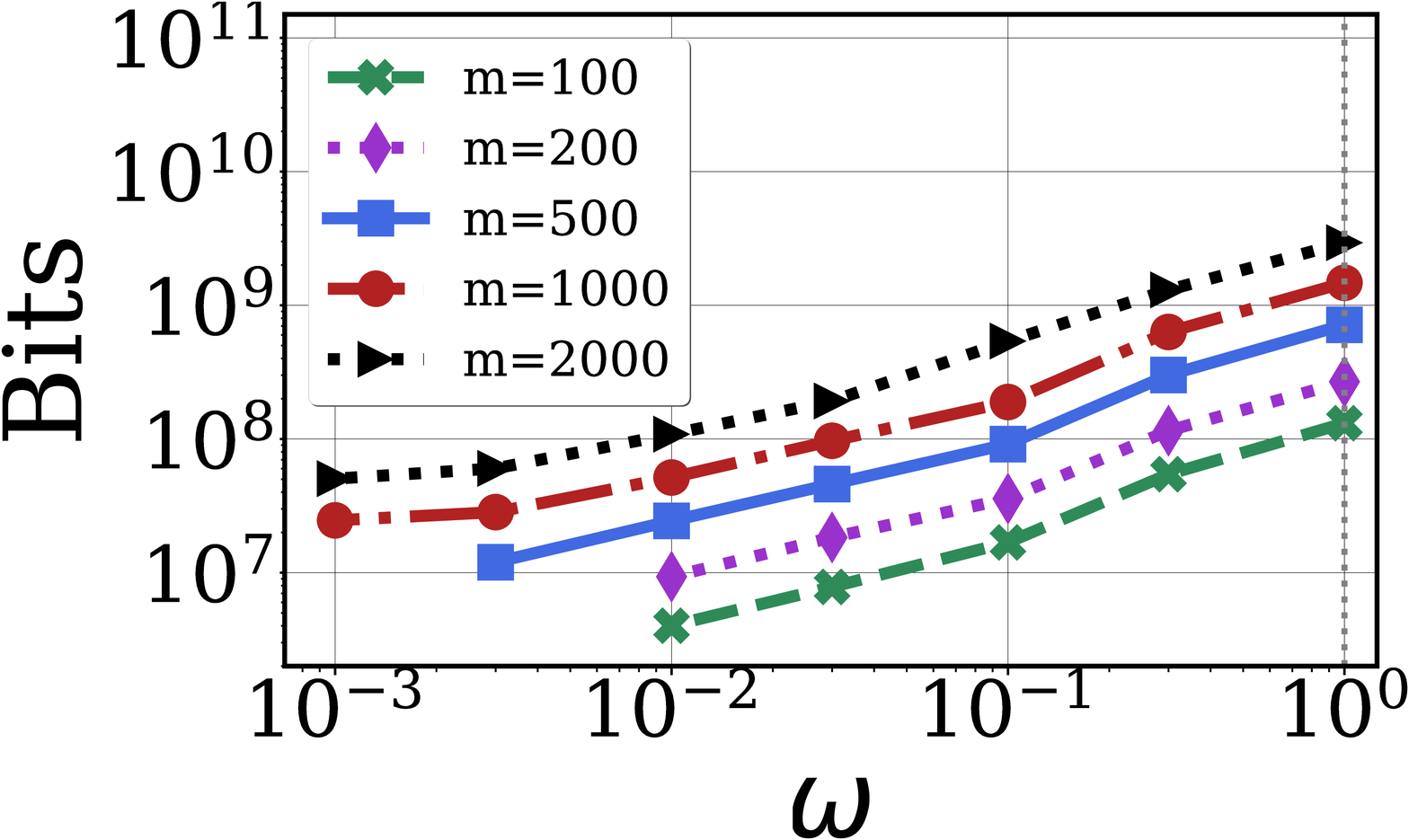}
    \subcaption{Path}
    \label{fig:m-top-omega-path}
    \end{minipage}
    \hspace{0.2em}
    \begin{minipage}{0.3\linewidth}
    \raggedleft
    \includegraphics[width=\linewidth]{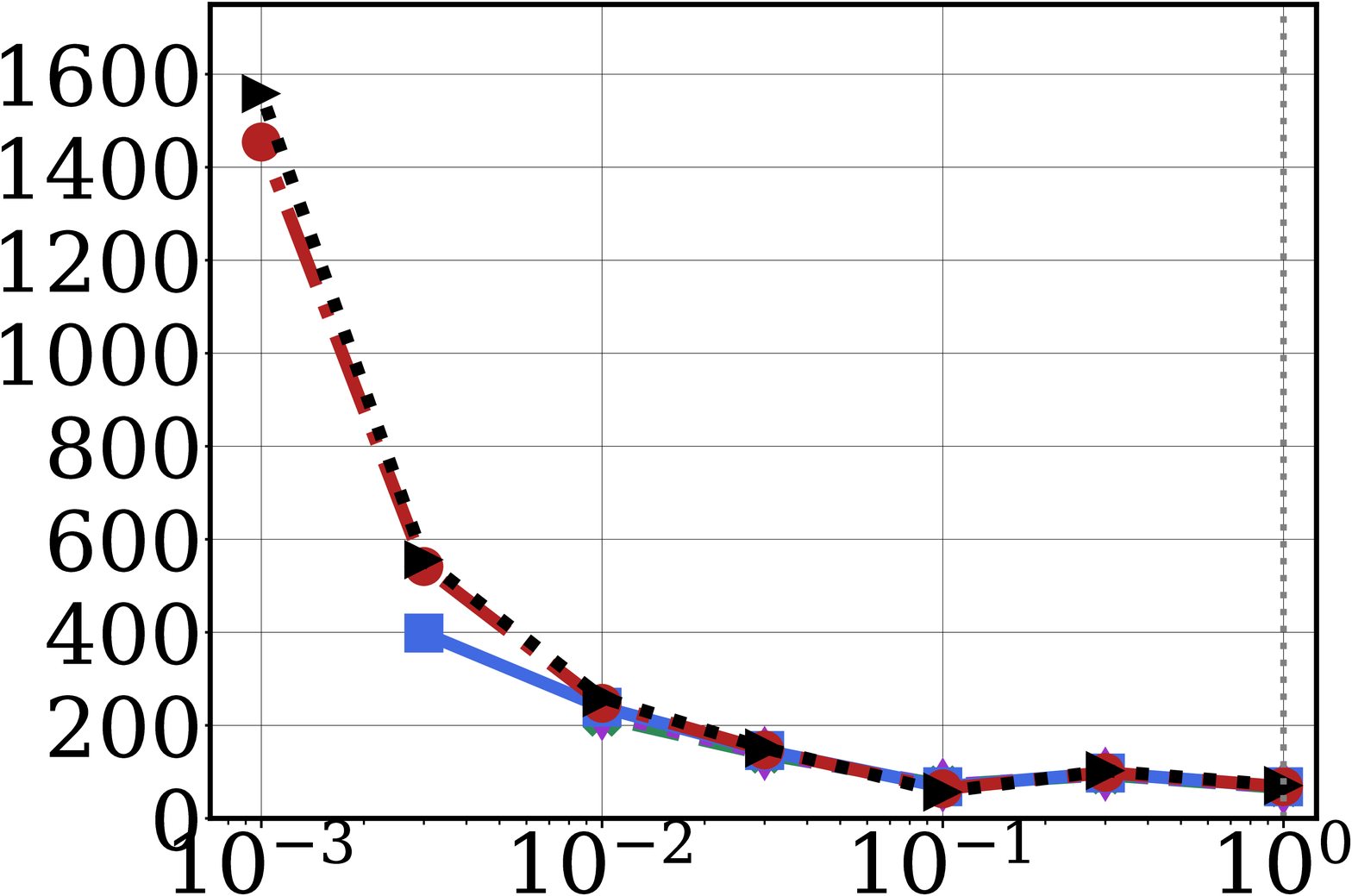}
    \includegraphics[width=\linewidth]{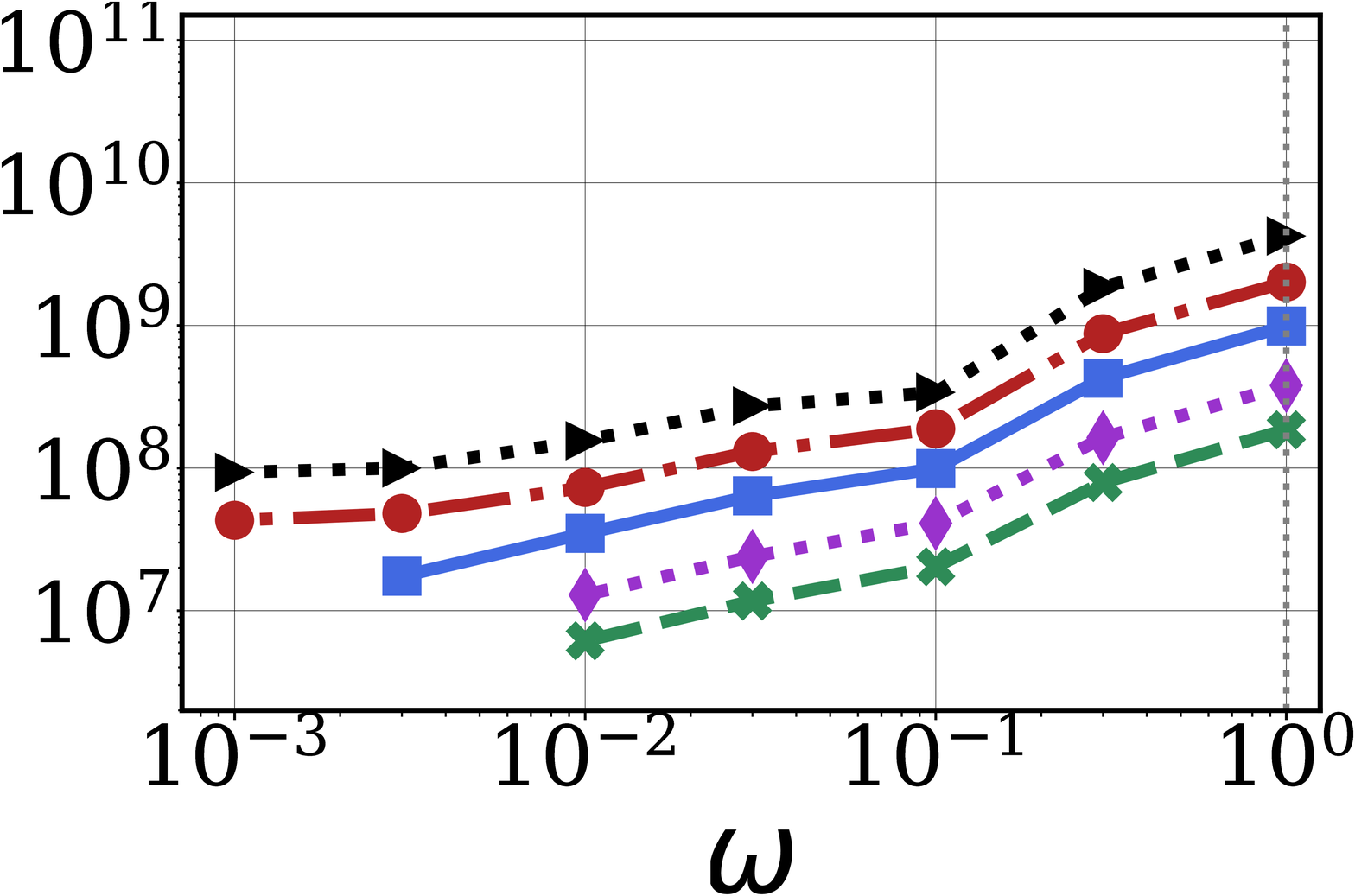}
    \subcaption{Torus}
    \label{fig:m-top-omega-torus}
    \end{minipage}
    \hspace{0.2em}
    \begin{minipage}{0.3\linewidth}
    \raggedleft
    \includegraphics[width=\linewidth]{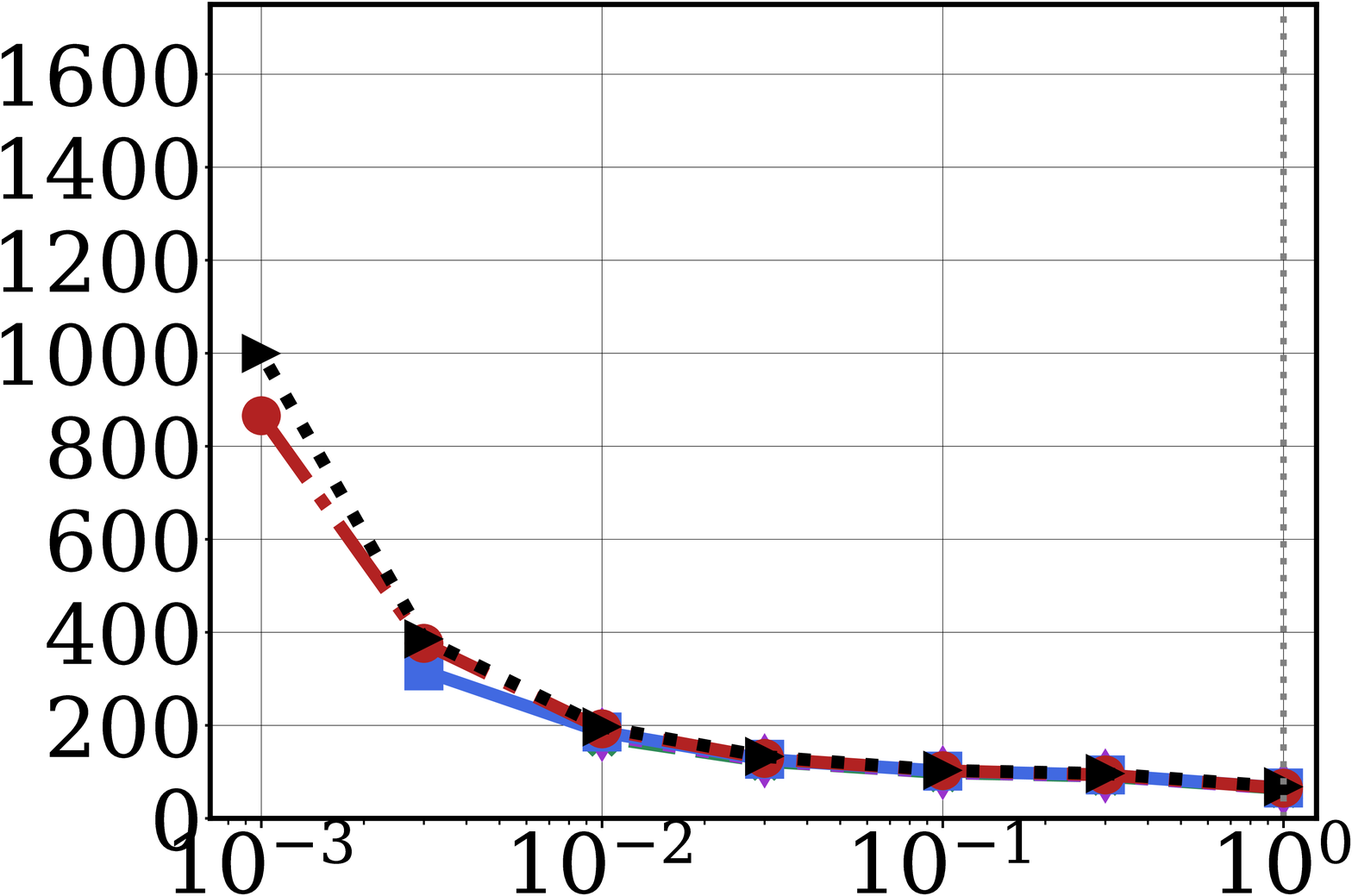}
    \includegraphics[width=\linewidth]{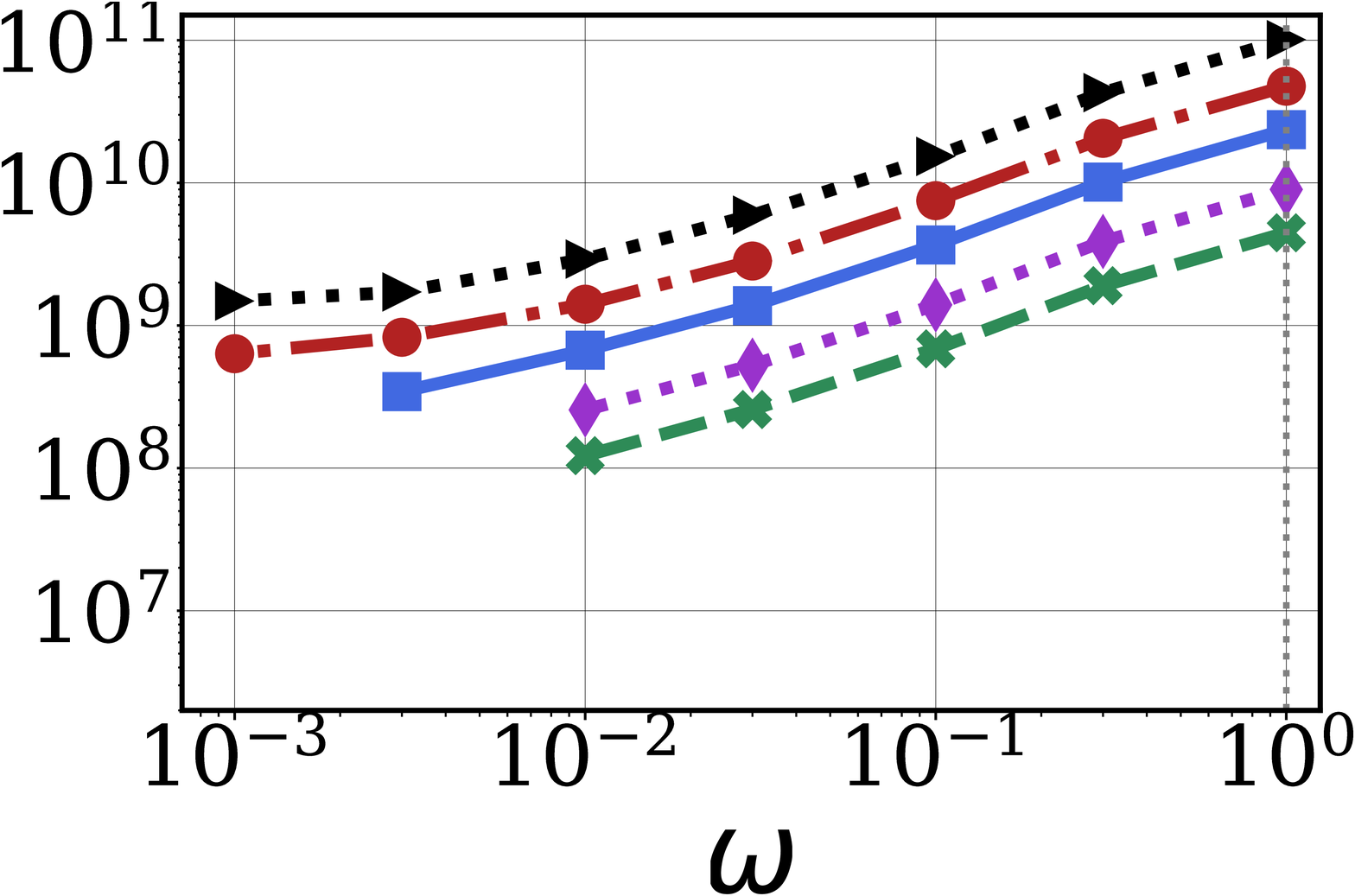}
    \subcaption{Complete}
    \label{fig:m-top-omega-complete}
    \end{minipage}
    \caption{
    The empirical mean over $100$ Monte Carlo runs of the number of iterations (top) and transmitted bits (bottom), required for \mbox{$\tilde{\mu}_i^t(\theta)<\epsilon$} for all agents on every $\theta\notin\Theta^\star$ with \mbox{$n=100$} agents and \mbox{$\mathrm{top}_{100\omega\%}$} operator over (a) path, (b) torus, and (c) complete topologies. Each line corresponds to a separate value of $m$ for different values of $\omega$.}
    \label{fig:m-top}
\end{figure}

\textbf{Networks Size:} We further quantify the mutual effect of compression ratio $\omega$ with the network size $n$, as well as the hypotheses size $m$ on the convergence rate in Fig.~\ref{fig:n-top} and Fig.~\ref{fig:m-top}, respectively. We first fix \mbox{$m=400$} and for different network sizes \mbox{$n\in\{25,50,100,200\}$}, we initialize beliefs uniformly and run $100$ Monte Carlo runs of Eq.~\eqref{eq:update-mu-alg} with compression operator \mbox{$\mathrm{top}_{100\omega\%}$} for \mbox{$\omega\in\{0.0025,0.005,0.01,0.025,0.05,0.1,0.2,0.5\}$}. For the choice of optimal $\gamma$ we apply a fine geometric grid search. 

In the top row of Fig.~\ref{fig:n-top}, we see the number of iterations required for agents to reach \mbox{$\epsilon=10^{-8}$} accuracy of the optimal hypothesis. In the bottom row, the number of bits required for the same experiments is shown.
For example, in Fig.~\ref{fig:n-top-omega-torus} regarding the torus topology, we can see that for different agent numbers $n$, the number of iterations required for consensus is very similar. In addition, the number of transmitted bits decays to \mbox{$2.5\%$} with \mbox{$\mathrm{top}_{1}$} communication (\mbox{$\omega=0.0025$}). More importantly, with \mbox{$\omega=0.1$}, the number of transmitted bits will be decreased to \mbox{$10\%$} of the \mbox{$\mathrm{full}$} communication with roughly the same number of iterations.

\textbf{Number of Hypotheses:} Next, we fix $n=100$ and for different number of hypotheses \mbox{$m\in\{100,200,500,100,2000\}$}, we run the same set of experiments. For each $m$, we consider the set of $\omega\geq {1}/{m}$. In the top and bottom rows of Fig.~\ref{fig:m-top}, the number of iterations and transmitted bits required for \mbox{$\epsilon$-convergence} (see Corollary~2, and the discussion on the convergence rate) are shown, respectively. In Fig.\ref{fig:m-top-omega-path}, we can for example see that for a path graph with $n=100$ agents and $m=2000$ hypotheses, the communication cost can be decreased to $2\%$.

\section{Conclusions And Future Work}
\label{sec:conclusion}
We proposed a distributed non-Bayesian update rule where agents exchange compressed messages with an arbitrary compression ratio to reach a consensus~\cite{koloskova2019decentralized}. Our algorithm leverages a unified compression mechanism that can embrace a wide range of quantization and sparsification operators. Our main results show that the beliefs generated by our proposed algorithm exponentially concentrates around the set of optimal hypotheses. Furthermore, we presented a probabilistic explicit convergence rate for our method. Finally, we presented empirical evidence suggesting that given a proper compression precision, our algorithm can reduce the required communication load compared to existing approaches (cf.~\cite{nedic2017fast}).

Our theoretical analyses and numerical results suggest a strong dependency of convergence rate on the topology of the network. In future work, we will explore how to reduce this dependency. Other aspects of future work would be to use social sampling in compression to provide more comprehensive models for agents' behavior in social networks. Future work should also study the convergence rates for the concentration of beliefs on time-varying and directed networks and robustness to stubborn agents. 

\bibliographystyle{IEEEbib}
\bibliography{ref}

\clearpage
\appendix
\subsection{Memory-Efficient Algorithm}\label{app:memory-efficient-alg}
Here, we present a memory-efficient version of Algorithm~\ref{alg:compressed-social-learning}. In summary, we suggest each agent $i\in[n]$ to keep the geometric average of its neighbors' beliefs with weights correspondent to matrix $\mA$. In other words, each agent $i$, instead of allocating a distinguished part of its memory to save vectors \mbox{$\hat\vmu_j^t$} (\mbox{for all $j\in[n]$} such that \mbox{$(i,j)\in\mcE$}), can simply keep a weighted geometric average of them. Algorithm~\ref{alg:memory-efficient-compressed-social-learning} shows the pseudo-code of a memory-efficient implementation.

\begin{algorithm}[H]
	\caption{Memory-Efficient Distributed Non-Bayesian Learning with Compressed Communication}
	\label{alg:memory-efficient-compressed-social-learning}
	\textbf{Input:} initial beliefs $\tilde\vmu_i^0\in\bbR^m$, mixing matrix $\mA$, compression ratio $\omega\in (0,1]$, and learning stepsize $\gamma\in (0,1]$ \\
	\textbf{Procedure :}
	\begin{algorithmic}[1]
		\STATE{$\hat{\vmu}_i^0 := \vect{1}_m$,  $\vc_i^0 := \vect{1}_m$, and $\vmu_i^0 := \tilde\vmu_i^0, \quad \text{for all } i\in [n]$}
		\FOR{$t$ \textbf{in} $0,\dots,T-1$, in parallel for all $i \in [n]$}
		\STATE{$\vq_i^t:= Q(\log\vmu^{t}_i-\log\hat{\vmu}^{t}_i)$}
		\FOR{$j\in[n]$ such that $\mA_{ij}>0$ (including $j=i$)}
		\STATE{Send $\vq_i^t$ and receive $\vq_j^t$}
		\ENDFOR
		\STATE{Observe $s_i^{t+1}$}
		\STATE{for all $\theta\in \Theta$:
		\\\quad \textcolor{gray}{(a)}\,\, $\hat{\mu}_i^{t+1}(\theta) = \hat{\mu}_i^{t}(\theta). \exp\left(q_i^t(\theta)\right)$
		\\\quad \textcolor{gray}{(b)}\,\,$c_i^{t+1}(\theta) = c_i^{t}(\theta). \displaystyle\prod_{j=1}^{n} \exp\left(q_j^t(\theta)\right)^{\mA_{ij}}$
		\\\quad \textcolor{gray}{(c)}\,\,$\mu^{t+1}_i(\theta) = \mu^t_i(\theta). \left(\frac{c_i^{t+1}(\theta)}{\hat{\mu}_i^{t+1}(\theta)}\right)^{\gamma}.\ell_i\left(s_i^{t+1}|\theta\right)$
		}
		\STATE{$\tilde{\vmu}^{t+1}_i = \frac{1}{\vect{1}^\top\vmu^{t+1}_i}\vmu^{t+1}_i$}
		\ENDFOR
	\end{algorithmic}
	\textbf{Output:} final beliefs $\tilde{\vmu}_i^T, \quad \text{for all } i\in[n]$
\end{algorithm}

\subsection{Proof of Lemma~\ref{lem:bounded-xi}}\label{app:bounded-xi}
\begin{proof}(Lemma~\ref{lem:bounded-xi})
Let \mbox{$\mcL_i^t(\theta)=\log\left({\ell_i\left(s_i^{t+1}|\theta\right)}/{\ell_i\left(s_i^{t}|\theta\right)}\right)$} be likelihood ratio of two consecutive observations. Also, we define vectors
\mbox{$\vxi^{t}(\theta) = [\xi_1^t(\theta),\xi_2^t(\theta),\dots,\xi_n^t(\theta)]^\top$} and \mbox{$\vmcL^{t}(\theta) = [\mcL_1^t(\theta),\mcL_2^t(\theta),\dots,\mcL_n^t(\theta)]^\top$}. According to Eq.~\eqref{eq:update-nu-equiv}:
\begingroup
\allowdisplaybreaks
\begin{equation*}
\begin{alignedat}{2}
\vxi^{t+1}(\theta) &= \mB\vxi^{t}(\theta) &&+\vmcL^{t}(\theta)\\
            &= \mB^t\vxi^{1}(\theta) &&+ \sum_{r=1}^{t} \mB^{t-r}\vmcL^{r}(\theta)\\
            &= \mB^t\vxi^{1}(\theta) &&+\sum_{r=1}^{t} \left(\mB^{t-r}-\frac{1}{n}\vect{1}\vect{1}^\top\right)\vmcL^{r}(\theta)\\
            & &&+ \frac{1}{n}\vect{1}\vect{1}^\top \sum_{r=1}^{t}\vmcL^{r}(\theta), \quad \forall\theta\in\Theta.
\end{alignedat}
\end{equation*}
\endgroup
First notice that by Eq.~\eqref{eq:update-nu-equiv}, the following relation holds:
\begin{align*}
    \xi_i^1(\theta) = \log\frac{\nu^1_i(\theta)}{\nu^0_i(\theta)} = \sum_{j=1}^{n}\mB_{ij}\log\frac{\nu_j^0(\theta)}{\nu^0_i(\theta)} + \ell_i\left(s_i^1|\theta\right),
\end{align*}
where for all $i,j\in[n]$, \mbox{$\left\lvert\log\left({\nu_i^0(\theta)}/{\nu_j^0(\theta)}\right)\right\rvert\leq\log\left({1}/{\alpha_1}\right)$} by Assumption~\ref{assump:init}\ref{assump:init-mu}. Similarly, for all \mbox{$t_1,t_2\geq 0$}, we know that \mbox{$\left\lvert\log\left({\ell_i\left(s_i^{t_1}|\theta\right)}/{\ell_i\left(s_i^{t_2}|\theta\right)}\right)\right\rvert\leq\log\left({1}/{\alpha_2}\right)$} by Assumption~\ref{assump:init}\ref{assump:init-ell}. Thus, we can infer that
\begin{align}\label{eq:xi-1-theta-bound}
\left\lVert \vxi^1(\theta)\right\rVert\leq \sqrt{n}\log\frac{1}{\alpha_1\alpha_2}\leq 2\sqrt{n}\log\frac{1}{\alpha},
\end{align}
where the second inequality holds since \mbox{$\alpha=\min\{\alpha_1,\alpha_2\}$}. Also, by definition of \mbox{$\mcL_i^t(\theta)$}, we have
\begin{align*}
\sum_{r=1}^{t}\mcL_i^{r}(\theta) = \sum_{r=1}^{t} \log\frac{\ell_i\left(s_i^{r+1}|\theta\right)}{\ell_i\left(s_i^{r}|\theta\right)}=\log\frac{\ell_i\left(s_i^{t+1}|\theta\right)}{\ell_i\left(s_i^{1}|\theta\right)},
\end{align*}
therefore, it holds that
\begin{align}\label{eq:L-r-theta-bound}
    \max\left\{\left\lVert\vmcL^{r}(\theta) \right\rVert,\left\lVert\sum_{r=1}^{t}\vmcL^{r}(\theta)\right\rVert\right\}\leq \sqrt{n}\log\frac{1}{\alpha}.
\end{align}
Further, recall that $\mB = (1-\gamma)\mI + \gamma\mA$. Then, the spectral gap of $\mB$ is equal to $\gamma\delta$, hence
\begin{align*}
    \left\lVert \mB^k - \frac{1}{n} \vect{1}\vect{1}^\top\right\rVert \leq (1-\gamma\delta)^k,
\end{align*}
which helps us to show that
\begingroup
\allowdisplaybreaks
\begin{align}\label{eq:power-k-doubly-stochastic-matrix}
\left\lVert\sum_{r=1}^{t}\left(\mB^{t-r}-\frac{1}{n}\vect{1}\vect{1}^\top\right)\vmcL^{r}(\theta)\right\rVert&\leq \sum_{r=1}^{t}\left(1-\gamma\delta\right)^{t-r}\left\lVert\vmcL^{r}(\theta)\right\rVert\leq\nonumber\\
\frac{\left(1-\left(1-\gamma\delta\right)^t\right)}{\gamma\delta}\sqrt{n}\log\frac{1}{\alpha_2} &\leq\frac{\sqrt{n}}{\gamma\delta}\log\frac{1}{\alpha}.
\end{align}
\endgroup

Using the triangle inequality along with referring to Eq.~\eqref{eq:xi-1-theta-bound}, Eq.~\eqref{eq:L-r-theta-bound}, and Eq.~\eqref{eq:power-k-doubly-stochastic-matrix}, we finally have the following result:
\begingroup
\allowdisplaybreaks
\begin{align*}
\left\lVert\vxi^{t+1}(\theta)\right\rVert\leq& \left\lVert\mB^t\vxi^{1}(\theta)\right\rVert + \left\lVert\frac{1}{n}\vect{1}\vect{1}^\top \sum_{r=1}^{t}\vmcL^{r}(\theta)\right\rVert\\
+&\left\lVert\sum_{r=1}^{t} \left(\mB^{t-r}-\frac{1}{n}\vect{1}\vect{1}^\top\right)\vmcL^{r}(\theta)\right\rVert\\
\leq&\left\lVert\vxi^{1}(\theta)\right\rVert + \left\lVert\sum_{r=1}^{t}\vmcL^{r}(\theta)\right\rVert\\
+&\left\lVert\sum_{r=1}^{t} \left(\mB^{t-r}-\frac{1}{n}\vect{1}\vect{1}^\top\right)\vmcL^{r}(\theta)\right\rVert\\
\leq&\left(3+\frac{1}{\gamma\delta}\right)\sqrt{n}\log\frac{1}{\alpha}\leq\frac{4\sqrt{n}}{\gamma\delta}\log\frac{1}{\alpha},\quad\forall\theta\in\Theta,
\end{align*}
\endgroup
where by definition of $\vxi_i^t$ and $\vxi^t(\theta_k)$, 
\begin{align*}
    \sum_{i=1}^n\left\lVert\vxi_i^t\right\rVert^2 =  \sum_{k=1}^m\left\lVert\vxi^t(\theta_k)\right\rVert\leq \frac{16nm}{\gamma^2\delta^2}\left(\log\alpha\right)^2,
\end{align*}
which is constant and does not depend on $t$.
\end{proof}

\subsection{Proof of Lemma~\ref{lem:modified-choco}}\label{app:modified-choco-proof}

\begin{proof}(Lemma~\ref{lem:modified-choco})
From~\cite[Lemma~17]{koloskova2019decentralized}, we know that

\begingroup
\allowdisplaybreaks
\begin{align}\label{eq:choco-ineq-1}
    \left\lVert\mX^{t+1}-\overline{\mX} \right\rVert_F^2 \leq \left(1-\gamma\delta\right)^2\left(1+\tau_1\right)\left\lVert\mX^{t}-\overline{\mX} \right\rVert_F^2&\nonumber\\
    + \gamma^2\left(1+\tau_1^{-1}\right)\beta^2 \left\lVert\mX^t-\hat{\mX}^{t+1}\right\rVert_F^2&,
\end{align}
\endgroup
so, it is enough to modify~\cite[Lemma~18]{koloskova2019decentralized} as follows:
\begingroup
\allowdisplaybreaks
\begin{align}\label{eq:choco-ineq-2}
\bbE_{\vzeta}\left[\left\lVert\mX^{t+1}-\hat{\mX}^{t+2} \right\rVert_F^2\right] &\leq
\left(1-\omega\right)\left\lVert\mX^{t+1}-\hat{\mX}^{t+1}+ \mZ^{t+1}\right\rVert_F^2\nonumber \\
&\leq\left(1+\tau_0\right)\left(1-\omega\right)\left\lVert\mX^{t+1}-\hat{\mX}^{t+1}\right\rVert_F^2\nonumber \\
&
\hspace{-2em}
+\left(1+\tau_0^{-1}\right)\left(1-\omega\right)\left\lVert\mZ^{t+1}\right\rVert_F^2,
\end{align}
\endgroup
where the second inequality follows~\cite[Remark~9]{koloskova2019decentralized}. Note that \mbox{$\bbE_{\vzeta}[.]$} appears when $Q(.)$ is a randomized compression operator, however, if we assume a deterministic operator, we can simply drop the expectation. Finally, according to the analysis for~\cite[Lemma~18]{koloskova2019decentralized}, we know that:
\begingroup
\allowdisplaybreaks
\begin{align}\label{eq:choco-ineq-3}
    \left\lVert\mX^{t+1}-\hat{\mX}^{t+1}\right\rVert_F^2 &\leq\gamma^2\beta^2\left(1+\tau_2^{-1}\right)\left\lVert\mX^t-\overline{\mX}\right\rVert_F^2\nonumber\\
    &
    \hspace{-2em}
    +\left(1+\gamma\beta^2\right)\left(1+\tau_2\right)\left\lVert\mX^t-\hat{\mX}^{t+1}\right\rVert_F^2.
\end{align}
\endgroup

By definition of $e_t$ and Eq.~\eqref{eq:choco-ineq-1}, Eq.~\eqref{eq:choco-ineq-2}, and Eq.~\eqref{eq:choco-ineq-3}, we have
\begin{align*}
\begin{split}
e_{t+1} &\leq U(\gamma)\left\lVert\mX^t-\overline{\mX}\right\rVert_F^2\\
& + V(\gamma)\left\lVert\mX^t-\hat{\mX}^{t+1}\right\rVert_F^2\\
& + L\left\lVert\mZ^{t+1}\right\rVert_F^2
\\
&\leq \max\{U(\gamma),V(\gamma)\} e_{t} + L z_{t},
\end{split}
\end{align*}
where
\begin{align}\label{eq:eta-parameter}
\begin{split}
U(\gamma) &= \left(1-\gamma\delta\right)^2 \left(1+\tau_1\right)\\
&+ \left(1+\tau_0\right)\left(1-\omega\right)\gamma^2\beta^2 \left(1+\tau_2^{-1}\right),
\\
V(\gamma) &= \gamma^2\beta^2\left(1+\tau_1^{-1}\right)\\
&+ \left(1+\tau_0\right)\left(1-\omega\right)\left(1+\gamma\beta\right)^2\left(1+\tau_2\right),
\\
L &= \left(1+\tau_0^{-1}\right)\left(1-\omega\right),
\end{split}
\end{align}
thus it is enough to select \mbox{$\gamma,\tau_1,\tau_2,\tau_0$} such that there exists a constant $\eta<1$, such that
\begin{align*}
\max\{U(\gamma),V(\gamma)\}\leq\eta<1.
\end{align*}
So, let fix \mbox{$\tau_0\triangleq\frac{\omega}{2(1-\omega)}>0$}, where the fact that $\omega>0$, guarantees $\tau_0>0$. So, we can infer that
\begin{align*}
\left(1+\tau_0\right)\left(1-\omega\right) = 1-\frac{\omega}{2} = 1-\hat{\omega},
\end{align*}
where \mbox{$\hat{\omega} = \frac{\omega}{2}$}. Therefore, parameters in Eq.~\eqref{eq:eta-parameter} turns into the following setting:
\begingroup
\allowdisplaybreaks
\begin{align*}
U(\gamma) &= \left(1-\gamma\delta\right)^2 \left(1+\tau_1\right)\\
&+ \left(1-\hat{\omega}\right)\gamma^2\beta^2 \left(1+\tau_2^{-1}\right),\\
V(\gamma) &= \gamma^2\beta^2\left(1+\tau_1^{-1}\right)\\
&+\left(1-\hat{\omega}\right)\left(1+\gamma\beta\right)^2\left(1+\tau_2\right),\\
L &= \frac{(1-\omega)(2-\omega)}{\omega},
\end{align*}
\endgroup
thus, according to the proof of~\cite[Theorem~2]{koloskova2019decentralized}, for the choice of hyperparameters
\begingroup
\allowdisplaybreaks
\begin{align*}
\tau_0 &\triangleq\frac{\omega}{2(1-\omega)},\,\,\tau_1 \triangleq \frac{\gamma\delta}{2},\,\,\tau_2 \triangleq \frac{\hat{\omega}}{2} = \frac{\omega}{4},\\
\gamma^\star&\triangleq\frac{\delta^2\omega}{32\delta+2\delta^2+8\beta^2+4\delta\beta^2-8\delta\omega},
\end{align*}
\endgroup
the following statement holds:
\begin{align*}
\begin{split}
\max\{U(\gamma^\star),V(\gamma^\star)\}\leq 1-\frac{\delta^2\hat{\omega}}{82} = 1-\frac{\delta^2\omega}{164},
\end{split}
\end{align*}
thus $\eta\triangleq1-\frac{\delta^2\omega}{164}$.
\end{proof}

\subsection{Proof of Lemma~\ref{lem:var-range}}\label{app:var-range}

\begin{proof}(Lemma~\ref{lem:var-range})
We showed in Eq.~\eqref{eq:e-t-bound} that $e_t$ defined in Lemma~\ref{lem:modified-choco} is bounded. By definition of $e_t$, $\vx_i$, and the fact that $\overline{\vx}=\vect{0}$, the following inequality holds
\begingroup
\allowdisplaybreaks
\begin{align*}
    \bbE_{\vzeta}\left[ \left\lvert\log\mu_i^t(\theta)-\log\nu_i^t(\theta)\right\rvert^2\right]\leq&\bbE_{\vzeta}\left[\left\lVert\log\vmu_i^t-\log\vnu_i^t\right\rVert^2\right]\leq\\
    \sum_{i=1}^n\bbE_{\vzeta}\left[ \left\lVert\log\vmu_i^t-\log\vnu_i^t\right\rVert^2\right]
    \leq&e_t\leq \eta^t e_0 + \frac{LR^2(1-\eta^t)}{1-\eta},
\end{align*}
\endgroup
and by Jensen's inequality, we have
\begingroup
\allowdisplaybreaks
\begin{align*}
    \left\lvert\bbE_{\vzeta}\left[\log\mu_i^t(\theta)\right]-\log\nu_i^t(\theta)\right\rvert \leq& \bbE_{\vzeta}\left[ \left\lvert\log\mu_i^t(\theta)-\log\nu_i^t(\theta)\right\rvert\right]\\
    \leq&
    \hspace{-1.5em}
    {\left(\bbE_{\vzeta}\left[ \left\lvert\log\mu_i^t(\theta)-\log\nu_i^t(\theta)\right\rvert^2\right]\right)}^{\frac{1}{2}}.
\end{align*}
\endgroup

It is therefore enough to compute an upper bound for the right-hand side of Eq.~\eqref{eq:e-t-bound}. Further, \mbox{$\forall i\in[n]$, $\vmu_i^0 = \vnu_i^0$}, and \mbox{$\hat{\vmu}_i^0=\vect{1}$}, so
\begingroup
\allowdisplaybreaks
\begin{align*}
    e_0 &= \bbE_{\vzeta}\left[\sum_{i=1}^{n} \left\lVert \hat{\vx}_i^1\right\rVert^2\right] = \bbE_{\vzeta}\left[\sum_{i=1}^{n} \left\lVert \log\hat{\vmu}_i^1 - \log\hat{\vnu}_i^1\right\rVert^2\right] \\
    &= \bbE_{\vzeta}\left[\sum_{i=1}^{n} \left\lVert Q\left(\log\vmu_i^0\right)- \log\vnu_i^0\right\rVert^2\right]\\
    &\leq \left(1-\omega\right)\sum_{i=1}^n\left\lVert\log\vmu_i^0\right\rVert^2 \leq \left(1-\omega\right)nm\left(\log\alpha\right)^2.
\end{align*}
\endgroup

Finally, we can conclude that
\begingroup
\allowdisplaybreaks
\begin{align*}
\eta^t e_0 + \frac{LR^2(1-\eta^t)}{1-\eta}\leq e_0+\frac{LR^2}{1-\eta}&<\\
nm\left(\log\alpha\right)^2\left(1+\frac{2624(1-\omega)(2-\omega)}{\delta^4\gamma^2\omega^2}\right)&<\\
\frac{5249}{\delta^4\gamma^2\omega^2}nm\left(\log\alpha\right)^2&\Rightarrow\\
\left\lvert\bbE_{\vzeta}\left[\log\mu_i^t(\theta)\right]-\log\nu_i^t(\theta)\right\rvert \leq \frac{73\sqrt{nm}}{\delta^2\gamma\omega}\log\frac{1}{\alpha}.
\end{align*}
\endgroup
\end{proof}

\end{document}